\documentclass{article}

    \PassOptionsToPackage{numbers, compress}{natbib}
 \usepackage[preprint]{neurips_2026}

\usepackage[utf8]{inputenc} 
\usepackage[T1]{fontenc}    
\usepackage{hyperref}       
\usepackage{url}            
\usepackage{booktabs}       
\usepackage{amsfonts}       
\usepackage{nicefrac}       
\usepackage{microtype}      
\usepackage{xcolor}         
\usepackage{graphicx}       
\definecolor{ablationred}{RGB}{150, 35, 35}
\definecolor{ablationgreen}{RGB}{25, 115, 65}
\usepackage{amsmath}
\usepackage{float}
\usepackage{booktabs}   
\usepackage{multirow}   
\usepackage{colortbl}

\usepackage{authblk}

\title{Focusable Monocular Depth Estimation}

\author[1,2,*]{Yuxin Du}
\author[1,*]{Tao Lin}
\author[1]{Zile Zhong}
\author[1]{Runting Li}
\author[1]{Xiyao Chen}
\author[1]{Jiting Liu}
\author[2]{Chenglin Liu}
\author[2]{Ying-Cong Chen}
\author[3]{Yuqian Fu}
\author[1,+]{Bo Zhao}

\affil[1]{School of Artificial Intelligence, Shanghai Jiao Tong University}
\affil[2]{The Hong Kong University of Science and Technology (Guangzhou)}
\affil[3]{King Abdullah University of Science and Technology}
\affil[*]{These authors contributed equally to this work.}
\affil[+]{Corresponding author. Email: bo.zhao@sjtu.edu.cn}

\begin{document}

\maketitle

\begin{abstract}
  Monocular depth foundation models generalize well across scenes, yet they are typically optimized with uniform pixel-wise objectives that do not distinguish user-specified or task-relevant target regions from the surrounding context. We therefore introduce \textbf{F}ocusable Monocular \textbf{D}epth \textbf{E}stimation (FDE), a region-aware depth estimation task in which, given a specified target region, the model is required to prioritize foreground
  depth accuracy, preserve sharp boundary transitions, and maintain coherent global scene geometry. 
  To prioritize task-critical region modeling, we propose \textbf{FocusDepth}, a prompt-conditioned monocular relative depth estimation framework that guides depth modeling to focus on target regions via box/text prompts.
  The core \textbf{Multi-Scale Spatial-Aligned Fusion (MSSA)} in FocusDepth spatially aligns multi-scale features from Segment Anything Model 3 to the Depth Anything family and injects them through scale-specific, gated conditional fusion. This enables dense prompt cue injection without disrupting geometric representations, thereby endowing the depth estimation model with focused perception capability.
  To study FDE, we establish \textbf{FDE-Bench}, a target-centric monocular relative depth benchmark built from image--target--depth triplets across five datasets, containing 252.9K/72.5K train/val triplets and 972 categories spanning real-world and embodied simulation environments. 
  On FDE-Bench, FocusDepth consistently improves over globally fine-tuned DA2/DA3 baselines under both box and text       
  prompts, with the largest gains appearing in target boundary and foreground regions while preserving global scene geometry. Ablations show that MSSA's spatial alignment is the key design factor, as disrupting prompt--geometry
  correspondence increases AbsRel by up to 13.8\%.
\end{abstract}

\begin{figure}[t]
\centering
\includegraphics[width=0.8\linewidth]{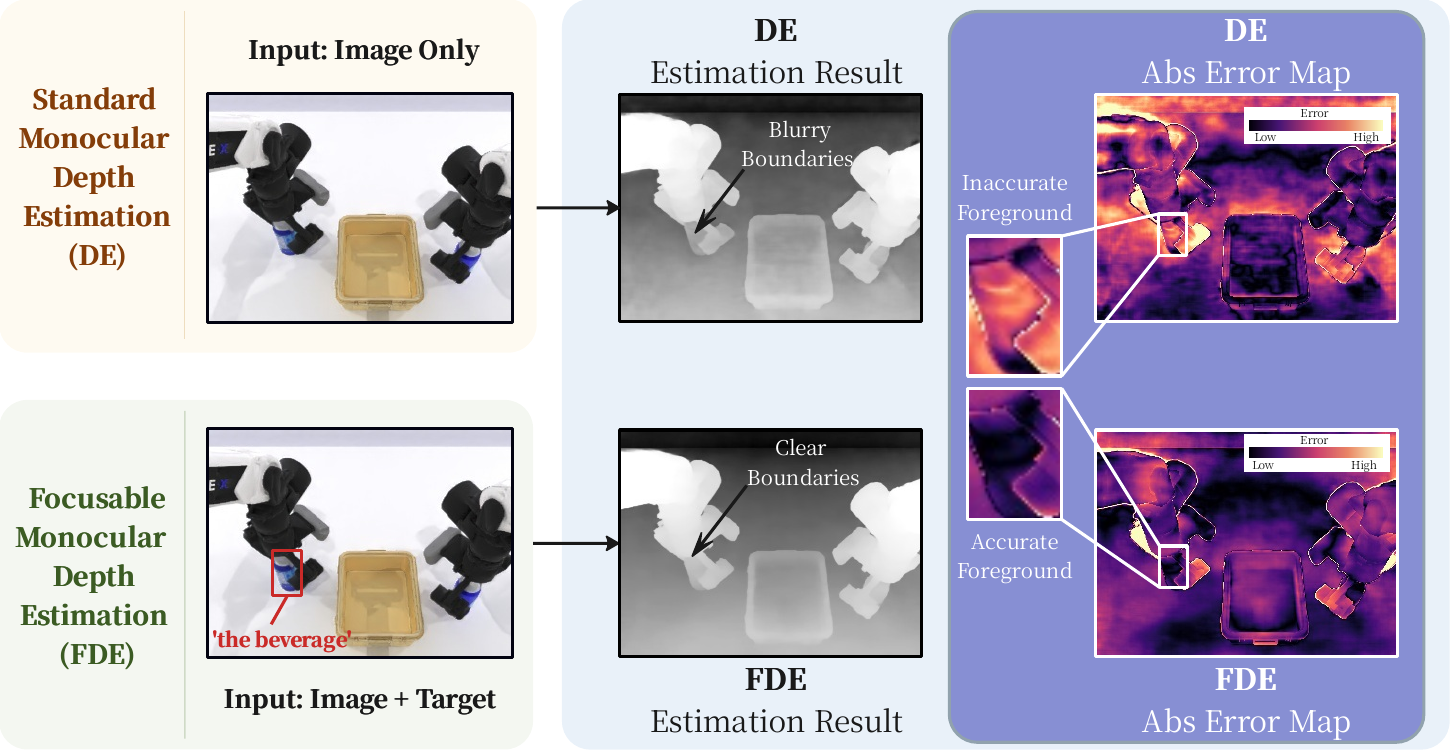}
\caption{Illustration of the proposed Focusable Depth Estimation (FDE). Given an image and a user-specified target, FDE prioritizes foreground and boundary depth quality while preserving global scene geometry. Compared with standard monocular depth estimation (DE), FDE requires the model to yield sharper target boundaries and lower depth error in the specified target region.}
\label{fig:concept}
\end{figure}

\section{Introduction}

Monocular depth estimation~\cite{ming2021deep, bhoi2019monocular, arampatzakis2023monocular} aims to recover dense scene geometry from a single RGB image and serves as a fundamental visual capability in many vision systems, including robotics, embodied AI, and augmented reality. The field has been extensively studied under supervised and self-supervised learning paradigms~\cite{zhao2020monocular,fu2018deep,godard2017unsupervised,godard2019digging}, with later work improving robustness through cross-dataset training, stronger architectures, and generative priors~\cite{ranftl2020towards,zhao2022monovit,zhang2023lite,ke2024repurposing,gui2025depthfm, zhang2025egonight}. More recently, depth foundation models such as Depth Anything~\cite{yang2024depth, yang2024depthv2, lin2025depth}, UniDepth~\cite{piccinelli2024unidepth, piccinelli2025unidepthv2}, and Metric3D~\cite{yin2023metric3d,hu2024metric3d} have substantially advanced cross-domain generalization by leveraging large-scale and diverse training data. Despite this progress, monocular depth estimation remains predominantly image-centric: models are typically trained and evaluated by aggregating errors over all pixels in the image. 
This uniform, image-wide treatment contrasts with task-driven spatial perception: humans selectively attend to goal-relevant foreground objects while retaining only contextual awareness of the background. This mismatch is also relevant to target-centric perception, with robot manipulation as one motivating example~\cite{black2024pi_0, intelligence2025pi_, lin2025evo0, lin2025evo1, wang2026oflow, wang2026ocra}, where depth near the target surface and discontinuities may be important but are not evaluated as downstream task performance here. As a result, strong full-image depth performance does not necessarily imply reliable target-local geometry under region-aware evaluation. Prior task-aware depth estimation has similarly observed that foreground objects can be underweighted by uniform image-wide objectives in downstream 3D perception~\cite{wang2020task}.


To address this formulation gap, we introduce \textbf{Focusable Depth Estimation (FDE)}, a short version of Focusable Monocular Depth Estimation for convenience, which extends monocular depth estimation from uniform dense prediction to target-conditioned depth inference. As illustrated in Figure~\ref{fig:concept}, given an image and a user-specified target region, FDE prioritizes target-foreground accuracy and boundary fidelity while preserving coherent global scene geometry.

To study FDE, we propose \textbf{FocusDepth}, a prompt-conditioned monocular relative depth estimation framework designed to balance local target sensitivity with global geometric coherence. Our key intuition is to combine the complementary priors of promptable segmentation and monocular depth foundation models: Segment Anything Model 3 (SAM3)~\cite{carion2025sam} provides prompt-grounded spatial selectivity for identifying user-specified target regions, while the Depth Anything family (DAs), such as DA2~\cite{yang2024depthv2} and DA3~\cite{lin2025depth}, provides a strong pretrained prior over dense scene geometry. However, directly fusing these models is nontrivial, as SAM3 is optimized for 2D prompt-driven localization whereas DAs are optimized for globally coherent 3D geometry. FocusDepth, therefore, treats target-aware refinement as a controlled injection problem rather than simple feature fusion. We introduce \textbf{MSSA} (\textbf{M}ulti-\textbf{S}cale \textbf{S}patial-\textbf{A}ligned \textbf{F}usion), which injects prompt-conditioned cues into DAs geometry tokens through spatially aligned token-level fusion, adapts them across depth feature scales, and uses gated conditional fusion to control prompt-conditioned corrections. This design encourages target-relevant regions, including foreground and boundary areas, to receive focused refinement while preserving DAs' full-scene geometric prior.

To make FDE measurable, we further establish \textbf{FDE-Bench}, a benchmark suite and evaluation protocol that reformulates existing depth datasets into image--target--depth triplets. Unlike conventional depth benchmarks that primarily report full-image error, FDE-Bench evaluates whether a method improves target-local geometry through three complementary regions: the specified foreground, the target boundary, and the full scene as a coherence constraint. The benchmark covers diverse RGB-D sources, including indoor scenes, tabletop object data, and simulated embodied-scene observations from NYU v2~\cite{silberman2012indoor}, TUM RGB-D~\cite{sturm2012benchmark}, YCB-Video~\cite{xiang2017posecnn}, RLBench~\cite{james2020rlbench}, and RoboTwin~\cite{mu2025robotwin}, containing 252.9K/72.5K train/val triplets. This protocol enables systematic comparison of standard monocular depth models and prompt-conditioned methods under target-centric criteria, assessing foreground accuracy, boundary fidelity, and global geometric consistency.

Our contributions are threefold:
\begin{itemize}
    \item We introduce Focusable Depth Estimation (FDE), a target-centric formulation of monocular depth estimation that prioritizes foreground accuracy and boundary fidelity for a user-specified region while preserving coherent global scene geometry.
    
    \item We propose FocusDepth, a prompt-conditioned monocular relative depth estimation framework that transfers prompt-grounded target evidence from SAM3 into DAs' dense geometry representation via MSSA, enabling spatially aligned, scale-dependent, and gated conditional depth refinement.
    
    \item We establish FDE-Bench, a benchmark suite and evaluation protocol that converts existing depth datasets into image--target--depth triplets and demonstrates that FocusDepth improves target-region accuracy and boundary fidelity while maintaining full-image geometric coherence.
\end{itemize}

\section{Related Work}
\label{sec:related_work}

\paragraph{Monocular depth estimation and depth foundation models.}
Monocular depth estimation has evolved from supervised and self-supervised pipelines to more transferable architectures trained across diverse datasets~\cite{fu2018deep,godard2017unsupervised,godard2019digging,ranftl2020towards,zhao2022monovit,zhang2023lite}. These developments have progressively improved depth accuracy, cross-domain robustness, and dense geometric reasoning, laying the groundwork for recent depth foundation models. Current leading models include UniDepth~\cite{piccinelli2024unidepth} and UniDepthV2~\cite{piccinelli2025unidepthv2}, Metric3D~\cite{yin2023metric3d} and Metric3Dv2~\cite{hu2024metric3d}, and the Depth Anything family~\cite{yang2024depth,yang2024depthv2,lin2025depth}. By leveraging large-scale training data, metric supervision, synthetic data, or auto-annotated depth, these models achieve strong zero-shot generalization and robust whole-image depth prediction across diverse scenes. However, they are still primarily formulated and evaluated at the image level, where objectives and metrics aggregate errors over the full image plane.

\paragraph{Task-aware and conditioned depth estimation.}
FDE is related to task-aware, semantic-guided, and conditioned depth formulations, but differs in how the condition is specified and what depth quality is prioritized. ForeSeE separates foreground and background depth prediction for 3D object detection, showing that uniform objectives can underweight task-critical foreground objects~\cite{wang2020task}; however, it targets category- and task-specific foregrounds rather than arbitrary user-specified regions. BriGeS fuses depth and segmentation foundation models to improve generalized monocular depth estimation in complex scenes~\cite{ma2026bridging}, whereas FocusDepth uses prompt-grounded SAM3 cues for region-prioritized refinement instead of global image-level enhancement. Other conditioned or object-centric methods either rely on additional geometric inputs, such as sparse LiDAR prompts~\cite{lin2025prompting} or external geometric priors~\cite{wang2025depthprior}, or focus on object-level/amodal 3D understanding~\cite{li2025amodal,chen2025sam}. In contrast, FDE uses a box or text prompt to specify the visible target region whose foreground accuracy and boundary fidelity should be improved while preserving full-scene geometric coherence.

\section{Method}
\label{sec:method}

\subsection{Problem Formulation}
\label{sec:method_problem_formulation}

Given an input image $I \in \mathbb{R}^{H \times W \times 3}$ and an interactive prompt $p$, FocusDepth aims to predict a dense depth map that improves depth quality in the specified foreground region while preserving coherent scene geometry over the full image. Following Section~\ref{sec:benchmark}, we consider two prompt types: box prompts and text prompts that specify the target object or region. The model is formulated as follows: $\hat{D}, \hat{M} = \mathcal{F}(I, p),$
where $\hat{D} \in \mathbb{R}^{H \times W}$ is the predicted depth map and $\hat{M} \in [0,1]^{H \times W}$ is an auxiliary prompt-conditioned  predicted foreground mask.

The objective of focusable depth estimation is spatially heterogeneous. The foreground region should be estimated more accurately, the boundary region around the target should preserve sharper depth transitions, and the global region should remain geometrically coherent. FocusDepth addresses this objective with prompt-conditioned feature fusion and region-aware supervision.

\subsection{Overall Framework}

\begin{figure}[t]
  \centering
  \includegraphics[width=0.9\linewidth]{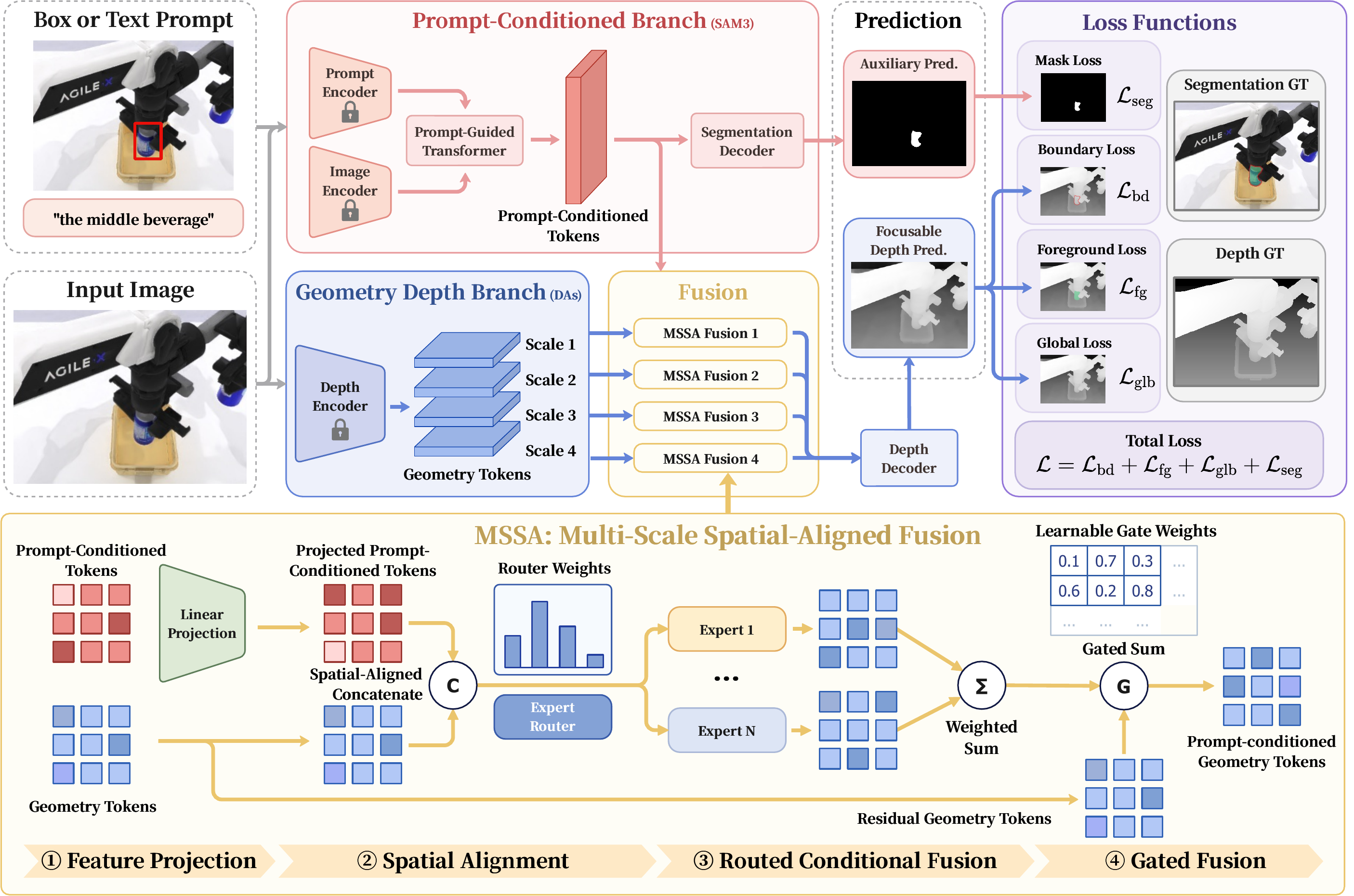}
  \caption{Overall framework of FocusDepth. The geometry depth branch extracts multi-scale geometry tokens, the prompt-conditioned branch produces prompt-conditioned tokens, and MSSA spatially aligns and injects target-aware cues into geometry representations for focusable depth prediction. The model is trained with foreground, boundary, and global depth losses, together with auxiliary segmentation supervision.}
  \label{fig:architecture}
  \end{figure}

  Figure~\ref{fig:architecture} illustrates the overall framework of FocusDepth. FocusDepth is built upon the Depth Anything family (DAs) as the geometry depth branch, including DA2~\cite{yang2024depthv2} and DA3~\cite{lin2025depth}, due to their strong zero-shot generalization and robust monocular geometric priors learned from large-scale depth data. SAM3~\cite{carion2025sam} is incorporated as the prompt-conditioned branch for target-aware region specification. The geometry depth branch extracts multi-scale geometry features, while the prompt-conditioned branch produces a prompt-conditioned token feature from the input image and prompt. The core module, MSSA, spatially aligns and fuses the prompt-conditioned tokens into each scale of geometry tokens for target-aware depth refinement.

  Given an input image $I$, the geometry depth branch extracts a hierarchy of encoder features $  X^g = \{X_s^{g}\}_{s=1}^{S}, X_s^{g} \in \mathbb{R}^{N \times C_g},$
  where $s$ indexes the feature scale. We use $S=4$ geometry to match the encoder scales. The prompt-conditioned branch takes the image and prompt $p$ as input and produces a prompt-conditioned token feature $X^{p} \in \mathbb{R}^{N \times C_p}.$
  The two branches share the same token count $N$ because they process the same image over an identical patch grid of ViT~\cite{dosovitskiy2020image}. FocusDepth applies MSSA at each geometry scale: $Y = \mathrm{MSSA}(X^g, X^{p}),$
  where each scale has independent fusion parameters. The fused features $Y = \{Y_s\}_{s=1}^{S}$ are passed to the depth decoder to predict $\hat{D}$. An auxiliary segmentation decoder on the prompt-conditioned branch predicts
  $\hat{M}$ during both training and inference.

\subsection{Multi-Scale Spatial-Aligned Fusion}

  MSSA is the central target-aware fusion module in FocusDepth. It injects prompt-conditioned cues into geometry tokens in a spatially adaptive manner while preserving
  the pretrained geometry prior. MSSA is applied independently at each geometry encoder scale.

  \paragraph{Scale-wise spatial alignment.}

  Under the shared setting of input image size 1008 and patch size 14 of ViT~\cite{dosovitskiy2020image}, the geometry depth branch and the prompt-conditioned branch produce image tokens on the same regular patch grid. Thus, their tokens are in one-to-one
  spatial correspondence. At scale $s$, the geometry depth branch provides geometry tokens $X_s^{g} \in \mathbb{R}^{N \times C_g}$, while the prompt-conditioned branch provides the prompt-conditioned tokens $X^{p} \in
  \mathbb{R}^{N \times C_p}$. We project $X^{p}$ into the scale-specific geometry-token space: $ \tilde{X}_s^{p} = X^{p} W_s^{p},$
  where $W_s^{p} \in \mathbb{R}^{C_p \times C_g}$ is a learnable projection. The projected prompt-conditioned tokens and geometry tokens are then concatenated token-wise: $Z_s = [X_s^{g} \,\|\, \tilde{X}_s^{p}] \in \mathbb{R}^{N \times 2C_g}.$
  This spatial-aligned concatenation allows prompt-conditioned target cues to guide refinement at the corresponding spatial locations at each geometry scale.

  \paragraph{Routed conditional fusion.}

  Different image locations can require different prompt-conditioned corrections. To provide conditional fusion capacity, MSSA applies a lightweight MoE layer to the fused tokens $Z_s$:
  $F_s = \mathrm{MoE}_s(Z_s) \in \mathbb{R}^{N \times C_g}.$
  We use $E=4$ experts at each scale. The router computes input-dependent combinations of expert outputs and is used as a flexible fusion module, not as an explicitly supervised decomposition into foreground, boundary, or background experts.

  \paragraph{Gated fusion output.}

  To avoid overwriting the geometry-token representation, MSSA combines the routed feature with the original geometry tokens through a learnable gate:
$  G_s = \sigma(F_s W_s^{g}),$
  where $W_s^{g} \in \mathbb{R}^{C_g \times 1}$ and $\sigma(\cdot)$ denotes the Sigmoid function. The output feature is $Y_s = G_s \odot F_s + (1-G_s) \odot X_s^{g}.$
  This gated formulation injects prompt-conditioned corrections selectively while retaining the pretrained geometry prior.

  \subsection{Learning Objective}
  \label{sec:method_region_loss}

  We use the foreground, boundary, and global region partition from FDE-Bench. For each training triplet, we follow the notation in Section~\ref{sec:benchmark}: $D$ is the
  ground-truth depth, $M$ is the foreground mask, $V$ is the valid-depth mask, and $V_{\mathrm{fg}}$, $V_{\mathrm{bd}}$, and $V_{\mathrm{glb}}$ denote the valid foreground,
  boundary, and global regions.

  In the relative depth setting, we align the prediction to the ground truth by solving a per-image scale $a$ and shift $b$ over valid pixels:
  $\tilde{D} = a\hat{D} + b.$
  Regional depth supervision is then computed as
  \begin{equation}
  \mathcal{L}_{\mathrm{fg}} = \mathcal{L}_{\mathrm{depth}}(\tilde{D}, D; V_{\mathrm{fg}}), \qquad
  \mathcal{L}_{\mathrm{bd}} = \mathcal{L}_{\mathrm{depth}}(\tilde{D}, D; V_{\mathrm{bd}}), \qquad
  \mathcal{L}_{\mathrm{glb}} = \mathcal{L}_{\mathrm{depth}}(\tilde{D}, D; V_{\mathrm{glb}}),
  \end{equation}
  where $\mathcal{L}_{\mathrm{depth}}$ combines masked MSE and gradient losses within the specified region. For DA2~\cite{yang2024depthv2}, we apply a disparity-space processing step before computing the loss.
    The auxiliary mask prediction is supervised by
  $
  \mathcal{L}_{\mathrm{seg}} =
  \mathcal{L}_{\mathrm{BCE}}(\hat{M}, M)
  +
  \mathcal{L}_{\mathrm{Dice}}(\hat{M}, M).
  $
  The final objective is
  $
  \mathcal{L} =
  \mathcal{L}_{\mathrm{glb}}
  +
  \mathcal{L}_{\mathrm{fg}}
  +
  \mathcal{L}_{\mathrm{bd}}
  +
  \mathcal{L}_{\mathrm{seg}}.
  $
  All terms are equally weighted because each loss is normalized within its valid region.
  
  \subsection{Two-Stage Training Strategy}
  \label{sec:method_training}

  To stabilize adaptation between the pretrained geometry and prompt-conditioned components, we train FocusDepth in two stages. In Stage 1, only MSSA is optimized, while all
  pretrained modules are frozen. This lets MSSA learn to align geometry tokens with prompt-conditioned tokens without disturbing the pretrained backbones.
  In Stage 2, we keep the depth encoder, the SAM3 image encoder, and the SAM3 prompt encoder frozen, and optimize MSSA, the depth decoder, the SAM3 prompt-guided transformer,
  and the auxiliary segmentation decoder. This separates cross-model alignment from downstream FDE adaptation and improves training stability.

\section{Proposed FDE-Bench}
\label{sec:benchmark}


\subsection{Benchmark Construction}

Existing monocular depth benchmarks are primarily image-centric: they evaluate depth quality over the whole image, but do not directly test whether a model can prioritize a user-specified target while preserving coherent geometry elsewhere. This mismatch is limiting for Focusable Depth Estimation (FDE), where the goal is to improve depth on the target region, preserve sharp depth transitions around it, and maintain globally consistent scene geometry. To make this setting operational, we establish FDE-Bench, a target-centric benchmark suite organized around the \emph{image--target--depth triplet}. Each triplet contains an RGB image, one target region, and the corresponding ground-truth depth map. A source image may yield multiple triplets, but each triplet is associated with exactly one target region, which is defined as the foreground region for evaluation. Targets may be specified through box prompts and, when available, text prompts.

FDE-Bench is constructed from heterogeneous RGB-D datasets by converting each retained target region into a standardized triplet. When source instance or semantic masks are available, annotated regions are filtered by area; otherwise, pseudo masks are generated using SAM automatic mask generation and filtered with the same protocol, with pseudo-mask entries explicitly marked in the metadata. Each retained mask defines the foreground region, its tight bounding box defines the box prompt, and its semantic class name, when available, defines the text prompt. We instantiate FDE-Bench on five adapted datasets: NYU v2~\cite{silberman2012indoor}, RLBench~\cite{james2020rlbench}, b~\cite{xiang2017posecnn}, TUM RGB-D~\cite{sturm2012benchmark}, and RoboTwin~\cite{mu2025robotwin}, covering indoor clutter, object-centric tabletop scenes, and embodied manipulation environments across real and simulated data. All entries derived from the same image or sequence are assigned to the same split to prevent data leakage. As illustrated in Appendix Table~\ref{tab:benchmark-stats}, FDE-Bench contains 20.5K/8.2K train/val images and 252.9K/72.5K train/val triplets, covering 972 target categories with semantic labels. Box-prompt evaluation is available across all five datasets, while text-prompt supervision is available on NYU v2, RLBench, and YCB-Video.

\subsection{Region-aware Evaluation Metrics}

FDE-Bench uses region-aware evaluation metrics aligned with the FDE objective. For each triplet, we evaluate three spatial regions: the foreground region, the boundary region, and the global region, respectively measuring target-local depth quality, depth-transition sharpness, and whole-image geometric consistency. Formally, let
$
D \in \mathbb{R}^{H \times W},
M \in \{0,1\}^{H \times W},
V \in \{0,1\}^{H \times W}
$
denote the ground-truth depth map, the target-region mask, and the valid-depth mask. We construct a boundary band $B$ around $M$ with radius $r=10$ pixels, given by the ring induced by morphological dilation and erosion. The valid foreground, boundary, and global regions are defined as
$
V_{\mathrm{fg}} = V \cap M,
V_{\mathrm{bd}} = V \cap B,
V_{\mathrm{glb}} = V.
$

For each region, we report $\boldsymbol{\delta_1}$ and \textbf{AbsRel}. Region metrics are computed in a per-target manner, which prevents images with many valid targets from dominating the final benchmark score. For relative-depth evaluation, scale-shift alignment is applied over the whole image before regional metrics are computed for coherent scene geometry alignment. More details of FDE-Bench, including dataset-specific construction rules, split protocols, prompt formats, representative visualizations, and additional statistics, are provided in Appendix~\ref{sec:app_benchmark}.

\section{Experiments}

\subsection{Experimental Setup}

We evaluate FocusDepth(DA2) and FocusDepth(DA3) on FDE-Bench using the dataset scope, prompt settings, and region-aware evaluation metrics defined in Section~\ref{sec:benchmark}. The framework design, region-aware learning objective, and training strategy follow Section~\ref{sec:method}. Unless otherwise specified, all experiments are conducted on 4 NVIDIA A100 80GB GPUs. We next describe the comparison settings and report quantitative and qualitative results.
To establish a baseline for the FDE capability of current representative monocular depth foundation models, we evaluate UniDepth-v1, UniDepth-v2, Metric3D-v2, Depth Anything 2 (DA2), and Depth Anything 3 (DA3) in the zero-shot setting. We further report the fine-tuning results of DA2 and DA3, and compare them with FocusDepth under the same training resources, data, and hyperparameter settings, detailed in Appendix Table~\ref{tab:focusdepth-training-hyperparams} and Appendix Table~\ref{tab:baseline-ft-training-hyperparams}.

\subsection{Main Results}
\label{sec:experiment_main_results}

Table~\ref{tab:main-box} and Table~\ref{tab:main-text} report the main box-prompt and text-prompt comparison on NYU v2, RLBench, RoboTwin, and YCB-Video. Appendix Table~\ref{tab:finetune-box-ycb-tum-diode} provides the remaining box-prompt results on TUM RGB-D and YCB-Video. Following the per-target evaluation protocol in FDE-Bench, we first compute metrics for each image--target--depth triplet and then report the median over target-level metrics, with Q25/Q75 shown for $\delta_1$. Across all tables, higher $\delta_1$ and lower AbsRel indicate better performance.

\begin{table*}[t]
\centering
\small
\setlength{\tabcolsep}{3.5pt}
\caption{Box-prompt results on RLBench, RoboTwin, and NYU v2. We compare zero-shot depth foundation models, fine-tuned DA2/DA3 baselines, and FocusDepth variants across boundary, foreground, and global regions.}
\label{tab:main-box}
\resizebox{0.9\textwidth}{!}{%
\begin{tabular}{llcccccc}
\toprule
\textbf{Dataset} & \textbf{Method} & \multicolumn{2}{c}{\textbf{Boundary region}} & \multicolumn{2}{c}{\textbf{Foreground region}} & \multicolumn{2}{c}{\textbf{Global region}} \\
\cmidrule(lr){3-4}\cmidrule(lr){5-6}\cmidrule(lr){7-8}
 & & $\delta_1$ $\uparrow$ & AbsRel $\downarrow$ & $\delta_1$ $\uparrow$ & AbsRel $\downarrow$ & $\delta_1$ $\uparrow$ & AbsRel $\downarrow$ \\

\midrule
\multirow{9}{*}{RLBench~\cite{james2020rlbench}} & UniDepth-v1~\cite{piccinelli2024unidepth} & 0.597 (0.386, 0.800) & 0.247 & 0.542 (0.008, 0.997) & 0.234 & 0.665 (0.574, 0.820) & 0.214 \\
 & UniDepth-v2~\cite{piccinelli2025unidepthv2} & \textbf{0.871} (\textbf{0.572}, \textbf{0.996}) & \textbf{0.123} & \textbf{0.997} (\textbf{0.355}, \textbf{1.000}) & \textbf{0.127} & \textbf{0.921} (\textbf{0.686}, \textbf{0.958}) & \textbf{0.097} \\
 & Metric3D-v2~\cite{hu2024metric3d} & 0.707 (0.470, 0.894) & 0.185 & 0.727 (0.081, \textbf{1.000}) & 0.198 & 0.815 (0.644, 0.903) & 0.165 \\
\cmidrule(lr){2-8}
 & DA2~\cite{yang2024depthv2} & 0.922 (0.680, 0.996) & 0.106 & 0.996 (0.694, 1.000) & 0.097 & 0.917 (0.726, 0.957) & 0.106 \\
 & \cellcolor{gray!10}DA2-ft & \cellcolor{gray!10}0.982 (0.889, 1.000) & \cellcolor{gray!10}0.069 & \cellcolor{gray!10}1.000 (0.949, 1.000) & \cellcolor{gray!10}0.066 & \cellcolor{gray!10}0.964 (0.953, 0.990) & \cellcolor{gray!10}0.054 \\
 & \cellcolor{gray!10}FocusDepth(DA2) & \cellcolor{gray!10}\textbf{0.996} (\textbf{0.960}, 1.000) & \cellcolor{gray!10}\textbf{0.054} & \cellcolor{gray!10}1.000 (\textbf{0.991}, 1.000) & \cellcolor{gray!10}\textbf{0.050} & \cellcolor{gray!10}\textbf{0.992} (\textbf{0.979}, \textbf{0.998}) & \cellcolor{gray!10}\textbf{0.045} \\
\cmidrule(lr){2-8}
 & DA3~\cite{lin2025depth} & 0.790 (0.488, 0.985) & 0.148 & 0.944 (0.103, 1.000) & 0.147 & 0.889 (0.648, 0.954) & 0.134 \\
 & \cellcolor{gray!10}DA3-ft & \cellcolor{gray!10}0.975 (0.804, 1.000) & \cellcolor{gray!10}0.073 & \cellcolor{gray!10}1.000 (0.709, 1.000) & \cellcolor{gray!10}0.095 & \cellcolor{gray!10}0.983 (0.921, 0.997) & \cellcolor{gray!10}0.042 \\
 & \cellcolor{gray!10}FocusDepth(DA3) & \cellcolor{gray!10}\textbf{0.996} (\textbf{0.949}, 1.000) & \cellcolor{gray!10}\textbf{0.049} & \cellcolor{gray!10}1.000 (\textbf{0.977}, 1.000) & \cellcolor{gray!10}\textbf{0.056} & \cellcolor{gray!10}\textbf{0.996} (\textbf{0.974}, \textbf{0.999}) & \cellcolor{gray!10}\textbf{0.030} \\
\midrule
\multirow{9}{*}{RoboTwin~\cite{mu2025robotwin}} & UniDepth-v1~\cite{piccinelli2024unidepth} & 0.751 (0.502, 0.929) & 0.190 & 0.840 (0.287, 0.995) & 0.151 & 0.930 (0.812, 0.975) & 0.102 \\
 & UniDepth-v2~\cite{piccinelli2025unidepthv2} & \textbf{0.919} (0.689, \textbf{0.987}) & \textbf{0.100} & \textbf{0.997} (\textbf{0.804}, \textbf{1.000}) & \textbf{0.073} & \textbf{0.983} (0.957, \textbf{0.993}) & 0.056 \\
 & Metric3D-v2~\cite{hu2024metric3d} & 0.905 (\textbf{0.703}, 0.979) & 0.101 & 0.988 (0.653, \textbf{1.000}) & 0.081 & \textbf{0.983} (\textbf{0.959}, 0.992) & \textbf{0.051} \\
\cmidrule(lr){2-8}
 & DA2~\cite{yang2024depthv2} & 0.954 (0.571, 0.991) & 0.095 & 0.993 (0.604, 1.000) & 0.082 & 0.984 (0.849, 0.997) & 0.062 \\
 & \cellcolor{gray!10}DA2-ft & \cellcolor{gray!10}0.957 (0.680, 0.995) & \cellcolor{gray!10}0.086 & \cellcolor{gray!10}0.987 (0.679, 1.000) & \cellcolor{gray!10}0.084 & \cellcolor{gray!10}0.989 (0.864, 0.997) & \cellcolor{gray!10}0.053 \\
 & \cellcolor{gray!10}FocusDepth(DA2) & \cellcolor{gray!10}\textbf{0.976} (\textbf{0.735}, \textbf{0.998}) & \cellcolor{gray!10}\textbf{0.070} & \cellcolor{gray!10}\textbf{0.995} (\textbf{0.878}, 1.000) & \cellcolor{gray!10}\textbf{0.065} & \cellcolor{gray!10}\textbf{0.994} (\textbf{0.880}, \textbf{0.999}) & \cellcolor{gray!10}\textbf{0.047} \\
\cmidrule(lr){2-8}
 & DA3~\cite{lin2025depth} & 0.889 (0.565, 0.987) & 0.119 & 0.973 (0.339, 1.000) & 0.115 & 0.962 (0.842, 0.989) & 0.077 \\
 & \cellcolor{gray!10}DA3-ft & \cellcolor{gray!10}0.949 (0.772, 0.995) & \cellcolor{gray!10}0.082 & \cellcolor{gray!10}0.992 (0.793, 1.000) & \cellcolor{gray!10}0.075 & \cellcolor{gray!10}0.985 (0.960, 0.997) & \cellcolor{gray!10}0.037 \\
 & \cellcolor{gray!10}FocusDepth(DA3) & \cellcolor{gray!10}\textbf{0.979} (\textbf{0.903}, \textbf{0.998}) & \cellcolor{gray!10}\textbf{0.058} & \cellcolor{gray!10}\textbf{0.998} (\textbf{0.955}, 1.000) & \cellcolor{gray!10}\textbf{0.055} & \cellcolor{gray!10}\textbf{0.995} (\textbf{0.977}, \textbf{0.999}) & \cellcolor{gray!10}\textbf{0.029} \\
 \midrule
\multirow{9}{*}{NYU v2~\cite{silberman2012indoor}} & UniDepth-v1~\cite{piccinelli2024unidepth} & \textbf{1.000} (\textbf{0.960}, 1.000) & 0.054 & 1.000 (\textbf{0.995}, 1.000) & 0.045 & 0.973 (0.945, 0.987) & 0.056 \\
 & UniDepth-v2~\cite{piccinelli2025unidepthv2} & 0.998 (0.948, 1.000) & \textbf{0.047} & 1.000 (0.993, 1.000) & \textbf{0.037} & \textbf{0.977} (\textbf{0.949}, \textbf{0.990}) & \textbf{0.046} \\
 & Metric3D-v2~\cite{hu2024metric3d} & 0.996 (0.916, 1.000) & 0.059 & 1.000 (0.979, 1.000) & 0.051 & 0.974 (0.928, \textbf{0.990}) & 0.059 \\
\cmidrule(lr){2-8}
 & DA2~\cite{yang2024depthv2} & 0.998 (0.946, 1.000) & 0.058 & 1.000 (0.991, 1.000) & 0.049 & 0.975 (\textbf{0.941}, 0.991) & \textbf{0.057} \\
 & \cellcolor{gray!10}DA2-ft & \cellcolor{gray!10}\textbf{1.000} (0.971, 1.000) & \cellcolor{gray!10}0.058 & \cellcolor{gray!10}1.000 (0.997, 1.000) & \cellcolor{gray!10}0.051 & \cellcolor{gray!10}\textbf{0.979} (0.938, \textbf{0.995}) & \cellcolor{gray!10}0.061 \\
 & \cellcolor{gray!10}FocusDepth(DA2) & \cellcolor{gray!10}\textbf{1.000} (\textbf{0.976}, 1.000) & \cellcolor{gray!10}\textbf{0.055} & \cellcolor{gray!10}1.000 (\textbf{0.998}, 1.000) & \cellcolor{gray!10}\textbf{0.048} & \cellcolor{gray!10}\textbf{0.979} (\textbf{0.941}, \textbf{0.995}) & \cellcolor{gray!10}0.059 \\
\cmidrule(lr){2-8}
 & DA3~\cite{lin2025depth} & 0.970 (0.869, 1.000) & 0.074 & 1.000 (0.986, 1.000) & 0.055 & 0.927 (0.857, 0.956) & 0.091 \\
 & \cellcolor{gray!10}DA3-ft & \cellcolor{gray!10}\textbf{1.000} (0.972, 1.000) & \cellcolor{gray!10}0.051 & \cellcolor{gray!10}1.000 (0.997, 1.000) & \cellcolor{gray!10}0.046 & \cellcolor{gray!10}0.977 (0.937, 0.993) & \cellcolor{gray!10}\textbf{0.053} \\
 & \cellcolor{gray!10}FocusDepth(DA3) & \cellcolor{gray!10}\textbf{1.000} (\textbf{0.977}, 1.000) & \cellcolor{gray!10}\textbf{0.049} & \cellcolor{gray!10}1.000 (\textbf{0.998}, 1.000) & \cellcolor{gray!10}\textbf{0.044} & \cellcolor{gray!10}\textbf{0.978} (\textbf{0.939}, \textbf{0.995}) & \cellcolor{gray!10}\textbf{0.053} \\
\bottomrule
\end{tabular}%
}
\end{table*}

\begin{table*}[t]
\centering
\small
\setlength{\tabcolsep}{3.5pt}
\caption{
Text-prompt results on RLBench, NYU v2, and YCB-Video. We compare zero-shot depth foundation models, fine-tuned DA2/DA3 baselines, and FocusDepth variants across boundary, foreground, and global regions.
}
\label{tab:main-text}
\resizebox{0.9\textwidth}{!}{%
\begin{tabular}{llcccccc}
\toprule
\textbf{Dataset} & \textbf{Method} & \multicolumn{2}{c}{\textbf{Boundary region}} & \multicolumn{2}{c}{\textbf{Foreground region}} & \multicolumn{2}{c}{\textbf{Global region}} \\
\cmidrule(lr){3-4}\cmidrule(lr){5-6}\cmidrule(lr){7-8}
 & & $\delta_1$ $\uparrow$ & AbsRel $\downarrow$ & $\delta_1$ $\uparrow$ & AbsRel $\downarrow$ & $\delta_1$ $\uparrow$ & AbsRel $\downarrow$ \\

\midrule
\multirow{9}{*}{RLBench~\cite{james2020rlbench}} & UniDepth-v1~\cite{piccinelli2024unidepth} & 0.601 (0.423, 0.775) & 0.247 & 0.577 (0.148, 0.958) & 0.224 & 0.666 (0.575, 0.822) & 0.212 \\
 & UniDepth-v2~\cite{piccinelli2025unidepthv2} & \textbf{0.859} (\textbf{0.579}, \textbf{0.990}) & \textbf{0.121} & \textbf{0.991} (\textbf{0.454}, \textbf{1.000}) & \textbf{0.117} & \textbf{0.925} (\textbf{0.695}, \textbf{0.958}) & \textbf{0.093} \\
 & Metric3D-v2~\cite{hu2024metric3d} & 0.716 (0.486, 0.870) & 0.183 & 0.753 (0.271, 0.998) & 0.185 & 0.817 (0.648, 0.905) & 0.163 \\
\cmidrule(lr){2-8}
 & DA2~\cite{yang2024depthv2} & 0.895 (0.691, 0.992) & 0.110 & 0.989 (0.781, 1.000) & 0.090 & 0.921 (0.728, 0.957) & 0.103 \\
 & \cellcolor{gray!10}DA2-ft & \cellcolor{gray!10}0.976 (0.888, 0.999) & \cellcolor{gray!10}0.069 & \cellcolor{gray!10}0.999 (0.949, 1.000) & \cellcolor{gray!10}0.061 & \cellcolor{gray!10}0.964 (0.953, 0.991) & \cellcolor{gray!10}0.054 \\
 & \cellcolor{gray!10}FocusDepth(DA2) & \cellcolor{gray!10}\textbf{0.994} (\textbf{0.961}, \textbf{1.000}) & \cellcolor{gray!10}\textbf{0.053} & \cellcolor{gray!10}\textbf{1.000} (\textbf{0.988}, 1.000) & \cellcolor{gray!10}\textbf{0.046} & \cellcolor{gray!10}\textbf{0.993} (\textbf{0.980}, \textbf{0.998}) & \cellcolor{gray!10}\textbf{0.044} \\
\cmidrule(lr){2-8}
 & DA3~\cite{lin2025depth} & 0.795 (0.507, 0.971) & 0.145 & 0.910 (0.294, 1.000) & 0.139 & 0.893 (0.655, 0.955) & 0.129 \\
 & \cellcolor{gray!10}DA3-ft & \cellcolor{gray!10}0.973 (0.824, 0.999) & \cellcolor{gray!10}0.068 & \cellcolor{gray!10}0.998 (0.772, 1.000) & \cellcolor{gray!10}0.088 & \cellcolor{gray!10}0.984 (0.923, 0.997) & \cellcolor{gray!10}0.041 \\
 & \cellcolor{gray!10}FocusDepth(DA3) & \cellcolor{gray!10}\textbf{0.995} (\textbf{0.959}, \textbf{1.000}) & \cellcolor{gray!10}\textbf{0.045} & \cellcolor{gray!10}\textbf{1.000} (\textbf{0.977}, 1.000) & \cellcolor{gray!10}\textbf{0.050} & \cellcolor{gray!10}\textbf{0.996} (\textbf{0.978}, \textbf{0.999}) & \cellcolor{gray!10}\textbf{0.029} \\
 \midrule
\multirow{9}{*}{NYU v2~\cite{silberman2012indoor}} & UniDepth-v1~\cite{piccinelli2024unidepth} & \textbf{0.998} (\textbf{0.952}, 1.000) & 0.054 & 1.000 (\textbf{0.991}, 1.000) & 0.046 & 0.973 (0.945, 0.987) & 0.056 \\
 & UniDepth-v2~\cite{piccinelli2025unidepthv2} & 0.994 (0.943, 1.000) & \textbf{0.048} & 1.000 (0.989, 1.000) & \textbf{0.038} & \textbf{0.977} (\textbf{0.949}, \textbf{0.990}) & \textbf{0.046} \\
 & Metric3D-v2~\cite{hu2024metric3d} & 0.990 (0.912, 1.000) & 0.060 & 1.000 (0.970, 1.000) & 0.052 & 0.974 (0.928, \textbf{0.990}) & 0.059 \\
\cmidrule(lr){2-8}
 & DA2~\cite{yang2024depthv2} & 0.995 (0.939, 1.000) & 0.059 & 1.000 (0.986, 1.000) & 0.050 & 0.975 (\textbf{0.941}, 0.991) & \textbf{0.057} \\
 & \cellcolor{gray!10}DA2-ft & \cellcolor{gray!10}\textbf{1.000} (0.960, 1.000) & \cellcolor{gray!10}0.059 & \cellcolor{gray!10}1.000 (0.993, 1.000) & \cellcolor{gray!10}0.052 & \cellcolor{gray!10}\textbf{0.979} (0.938, \textbf{0.995}) & \cellcolor{gray!10}0.061 \\
 & \cellcolor{gray!10}FocusDepth(DA2) & \cellcolor{gray!10}\textbf{1.000} (\textbf{0.963}, 1.000) & \cellcolor{gray!10}\textbf{0.057} & \cellcolor{gray!10}1.000 (\textbf{0.995}, 1.000) & \cellcolor{gray!10}\textbf{0.049} & \cellcolor{gray!10}\textbf{0.979} (0.940, \textbf{0.995}) & \cellcolor{gray!10}0.060 \\
\cmidrule(lr){2-8}
 & DA3~\cite{lin2025depth} & 0.959 (0.865, 1.000) & 0.077 & 1.000 (0.979, 1.000) & 0.057 & 0.927 (0.857, 0.956) & 0.091 \\
 & \cellcolor{gray!10}DA3-ft & \cellcolor{gray!10}\textbf{1.000} (0.961, 1.000) & \cellcolor{gray!10}0.052 & \cellcolor{gray!10}1.000 (0.992, 1.000) & \cellcolor{gray!10}0.047 & \cellcolor{gray!10}0.977 (0.937, 0.993) & \cellcolor{gray!10}\textbf{0.053} \\
 & \cellcolor{gray!10}FocusDepth(DA3) & \cellcolor{gray!10}\textbf{1.000} (\textbf{0.967}, 1.000) & \cellcolor{gray!10}\textbf{0.050} & \cellcolor{gray!10}1.000 (\textbf{0.995}, 1.000) & \cellcolor{gray!10}\textbf{0.044} & \cellcolor{gray!10}\textbf{0.978} (\textbf{0.938}, \textbf{0.995}) & \cellcolor{gray!10}0.054 \\
\midrule
\multirow{7}{*}{YCB-Video~\cite{xiang2017posecnn}} & UniDepth-v1~\cite{piccinelli2024unidepth} & \textbf{1.000} (0.985, 1.000) & 0.041 & 1.000 (0.997, 1.000) & 0.037 & 0.993 (0.981, 0.997) & 0.045 \\
 & UniDepth-v2~\cite{piccinelli2025unidepthv2} & 0.997 (0.966, 1.000) & \textbf{0.036} & 1.000 (0.994, 1.000) & \textbf{0.026} & 0.996 (\textbf{0.991}, 0.998) & \textbf{0.031} \\
 & Metric3D-v2~\cite{hu2024metric3d} & \textbf{1.000} (\textbf{0.986}, 1.000) & 0.041 & 1.000 (\textbf{0.998}, 1.000) & 0.031 & \textbf{0.997} (0.988, \textbf{0.999}) & 0.042 \\
 & DA2~\cite{yang2024depthv2} & 0.998 (0.963, 1.000) & 0.045 & 1.000 (0.993, 1.000) & 0.037 & 0.994 (0.972, 0.998) & 0.040 \\
\cmidrule(lr){2-8}
 & DA3~\cite{lin2025depth} & 0.993 (0.938, 1.000) & 0.053 & 0.999 (0.986, 1.000) & 0.044 & 0.992 (0.961, 0.997) & 0.047 \\
 & \cellcolor{gray!10}DA3-ft & \cellcolor{gray!10}\textbf{1.000} (0.988, 1.000) & \cellcolor{gray!10}0.042 & \cellcolor{gray!10}\textbf{1.000} (\textbf{0.999}, 1.000) & \cellcolor{gray!10}0.035 & \cellcolor{gray!10}0.997 (0.984, \textbf{0.999}) & \cellcolor{gray!10}0.044 \\
 & \cellcolor{gray!10}FocusDepth(DA3) & \cellcolor{gray!10}\textbf{1.000} (\textbf{0.990}, 1.000) & \cellcolor{gray!10}\textbf{0.034} & \cellcolor{gray!10}\textbf{1.000} (\textbf{0.999}, 1.000) & \cellcolor{gray!10}\textbf{0.024} & \cellcolor{gray!10}\textbf{0.998} (\textbf{0.993}, \textbf{0.999}) & \cellcolor{gray!10}\textbf{0.032} \\
\bottomrule
\end{tabular}%
}
\end{table*}

\paragraph{Zero-shot baselines.}
Table~\ref{tab:main-box} shows that current monocular depth foundation models already provide strong whole-image geometry in several settings, but their target-centric behavior is not consistently aligned with FDE. On NYU v2, zero-shot baselines are close to saturation in $\delta_1$, making differences mainly visible through AbsRel and lower-quartile statistics. On RLBench and RoboTwin, however, the gap between global performance and target-relevant boundary or foreground performance becomes more apparent. This suggests that strong image-centric depth estimation does not by itself ensure reliable focusable depth estimation on manipulation-oriented and object-centric scenes.

\paragraph{FocusDepth gains.}
Compared with the corresponding globally fine-tuned DA variants, FocusDepth more consistently improves the prompted target regions while preserving global geometry. The clearest gains appear on RLBench: under box prompts, FocusDepth(DA3) reduces boundary AbsRel from 0.073 to 0.049, foreground AbsRel from 0.095 to 0.056, and global AbsRel from 0.042 to 0.030 relative to DA3-ft. 
RoboTwin shows the same pattern, with FocusDepth(DA3) improving boundary, foreground, and global AbsRel over DA3-ft from 0.082/0.075/0.037 to 0.058/0.055/0.029. 
FocusDepth(DA3) mainly improves the boundary and foreground regions over DA3-ft on NYU v2, while matching DA3-ft in global AbsRel at 0.053, indicating that its target-region refinement does not compromise global geometric coherence.
FocusDepth(DA2) follows a similar trend over DA2-ft, indicating that the benefit is not tied to a single backbone. 
The text-prompt comparison in Table~\ref{tab:main-text} shows consistent gains on RLBench, NYU v2, and YCB-Video.
The appendix results further support this conclusion: the remaining box-prompt results in Table~\ref{tab:finetune-box-ycb-tum-diode} show clear improvements on TUM RGB-D and YCB-Video. Overall, these results support the FocusDepth objective: prompt-conditioned refinement most reliably improves local target regions, with limited global-geometry trade-off in most evaluated settings.

\subsection{RLBench Prompt-Correctness Study}
\label{sec:experiment_prompt_correctness}

We conduct this study to examine the sensitivity of FocusDepth(DA3) to prompt correctness in the FDE task. The evaluation includes a correct target prompt, a wrong prompt referring to another
visible object, a wrong prompt referring to an absent object, and an empty prompt. Unlike the main comparison, which reports medians, this study reports means to better expose the aggregate effect of bad cases under
prompt perturbations; the corresponding mean--median comparison is provided in Appendix Table~\ref{tab:app-prompt-correctness-mean-median}.

\begin{figure*}[h]
\centering
\begin{minipage}[t]{0.55\textwidth}
\centering
(a) Prompt-conditioned attention visualization.

\vspace{0.9em}
\includegraphics[width=\linewidth]{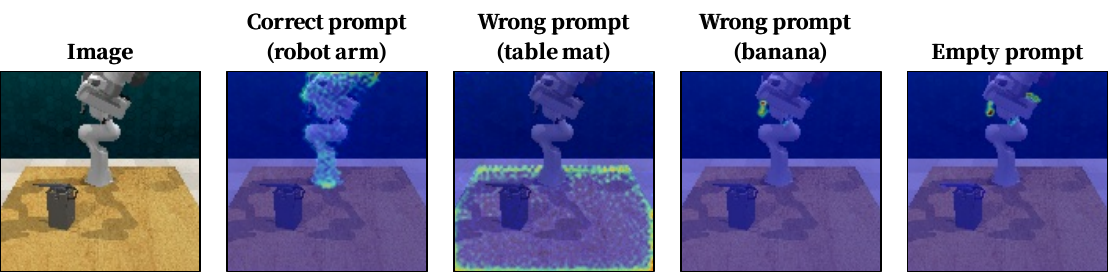}
\end{minipage}
\hfill
\begin{minipage}[t]{0.44\textwidth}
\centering
(b) Quantitative prompt-correctness study with mean values.

\vspace{0.3em}
\scriptsize
\setlength{\tabcolsep}{2.5pt}
\renewcommand{\arraystretch}{0.9}
\resizebox{0.9\linewidth}{!}{%
\begin{tabular}{@{}lcccccc@{}}
\toprule
Prompt Setting & \multicolumn{2}{c}{Boundary region} & \multicolumn{2}{c}{Foreground region} & \multicolumn{2}{c}{Global region} \\
\cmidrule(lr){2-3}\cmidrule(lr){4-5}\cmidrule(lr){6-7}
 & $\delta_1$ $\uparrow$ & AbsRel $\downarrow$ & $\delta_1$ $\uparrow$ & AbsRel $\downarrow$ & $\delta_1$ $\uparrow$ & AbsRel $\downarrow$ \\
\midrule
Correct & \textbf{0.919} & \textbf{0.108} & \textbf{0.901} & \textbf{0.137} & \textbf{0.943} & 0.075 \\
Wrong visible & 0.914 & 0.111 & 0.884 & 0.148 & \textbf{0.943} & \textbf{0.074} \\
Wrong absent & 0.911 & 0.117 & 0.886 & 0.152 & 0.939 & 0.078 \\
Empty & 0.910 & 0.118 & 0.886 & 0.154 & 0.939 & 0.078 \\
DA3-ft & 0.860 & 0.162 & 0.804 & 0.226 & 0.915 & 0.105 \\
\bottomrule
\end{tabular}%
}
\renewcommand{\arraystretch}{1.0}
\end{minipage}
\caption{RLBench prompt-correctness study for FocusDepth(DA3) under class-specific text prompts. (a) Prompt-conditioned attention visualization under correct, wrong-visible, absent, and empty
  prompts. (b) Quantitative results reported as means; the corresponding mean--median comparison is provided in Appendix Table~\ref{tab:app-prompt-correctness-mean-median}. Best results are shown in bold.}
\label{fig:rlbench-prompt-correctness}
\end{figure*}

As shown in Figure~\ref{fig:rlbench-prompt-correctness}, correct prompts produce the strongest target-centric response and the best boundary and foreground results. Degraded prompts mainly reduce local target-region
gains, with smaller effects on global geometry. Notably, even wrong or empty prompts remain stronger than the DA3-ft no-prompt baseline across boundary, foreground, and global regions. This suggests that, after
removing the benefit of correct prompt guidance, the prompt-conditioned branch still provides useful local and global depth cues.

\subsection{Ablation Studies}
\label{sec:experiment_ablation}

We ablate FocusDepth(DA3) on RLBench to examine how each design contributes to target-conditioned dense refinement. Table~\ref{tab:ablation-rlbench} reports median AbsRel under
box and text prompts across boundary, foreground, and global regions. The results show a clear hierarchy: spatial alignment is most critical, scale-specific fusion and
routed conditional fusion contribute to MSSA capacity, and gated injection, region-aware loss, and two-stage training stabilize the local-global trade-off.

The MSSA ablations identify spatial alignment as the dominant factor. Shuffling the correspondence between prompt-conditioned tokens and geometry tokens causes the largest degradation, increasing text-prompt global AbsRel by 13.8\%. 
Removing scale-specific fusion or replacing routed conditional fusion
with a single MLP also consistently worsens foreground and boundary AbsRel; for example, both variants increase text-prompt foreground AbsRel by 10.0\%, indicating that MSSA benefits from scale-aware and conditional fusion capacity for target-conditioned refinement.
Removing gated injection leads to smaller but consistent degradation, such as increasing box-prompt foreground AbsRel by 5.4\%, supporting its role in stabilizing prompt-conditioned corrections while preserving the pretrained geometry
prior.

The optimization ablations clarify the local-global trade-off. The global-loss-only variant improves global AbsRel but substantially worsens foreground AbsRel, e.g., by
10.7\% under box prompts and 12.0\% under text prompts, showing that image-centric supervision can sacrifice the prompted target region. One-stage training also weakens
local refinement, indicating that cross-model alignment and downstream FDE adaptation benefit from being separated. Together, these results support MSSA as a spatially
aligned, multi-scale, and conditional injection mechanism, and validate the region-aware objective and two-stage training strategy.

\begin{table*}[t]
\centering
\small
\setlength{\tabcolsep}{5pt}
\caption{Ablation study of FocusDepth(DA3) on RLBench. We report median AbsRel; lower is better. Colored values in parentheses denote changes relative to the full model.}
\label{tab:ablation-rlbench}
\resizebox{\textwidth}{!}{%
\begin{tabular}{lcccccc}
\toprule
Method & \multicolumn{3}{c}{Box prompt} & \multicolumn{3}{c}{Text prompt} \\
\cmidrule(lr){2-4}\cmidrule(lr){5-7}
 & Boundary & Foreground & Global & Boundary & Foreground & Global \\
\midrule
Full model (FocusDepth) & 0.049 & 0.056 & 0.030 & 0.045 & 0.050 & 0.029 \\
w/o spatial alignment (shuffled tokens) & 0.054 \textcolor{ablationred}{(+10.2\%)} & 0.062 \textcolor{ablationred}{(+10.7\%)} & 0.033 \textcolor{ablationred}{(+10.0\%)} & 0.050 \textcolor{ablationred}{(+11.1\%)} & 0.056 \textcolor{ablationred}{(+12.0\%)} & 0.033 \textcolor{ablationred}{(+13.8\%)} \\

w/o scale-specific fusion & 0.052 \textcolor{ablationred}{(+6.1\%)} & 0.060 \textcolor{ablationred}{(+7.1\%)} & 0.032 \textcolor{ablationred}{(+6.7\%)} & 0.048 \textcolor{ablationred}{(+6.7\%)} & 0.055 \textcolor{ablationred}{(+10.0\%)} & 0.031 \textcolor{ablationred}{(+6.9\%)} \\

w/o routed fusion (single MLP) & 0.052 \textcolor{ablationred}{(+6.1\%)} & 0.061 \textcolor{ablationred}{(+8.9\%)} & 0.032 \textcolor{ablationred}{(+6.7\%)} & 0.048 \textcolor{ablationred}{(+6.7\%)} & 0.055 \textcolor{ablationred}{(+10.0\%)} & 0.031 \textcolor{ablationred}{(+6.9\%)} \\

w/o gated injection & 0.050 \textcolor{ablationred}{(+2.0\%)} & 0.059 \textcolor{ablationred}{(+5.4\%)} & 0.031 \textcolor{ablationred}{(+3.3\%)} & 0.047 \textcolor{ablationred}{(+4.4\%)} & 0.051 \textcolor{ablationred}{(+2.0\%)} & 0.031 \textcolor{ablationred}{(+6.9\%)} \\

global loss only & 0.049 (+0.0\%) & 0.062 \textcolor{ablationred}{(+10.7\%)} & 0.028 \textcolor{ablationgreen}{(-6.7\%)} & 0.045 (+0.0\%) & 0.056 \textcolor{ablationred}{(+12.0\%)} & 0.028 \textcolor{ablationgreen}{(-3.4\%)} \\

one-stage training & 0.050 \textcolor{ablationred}{(+2.0\%)} & 0.059 \textcolor{ablationred}{(+5.4\%)} & 0.030 (+0.0\%) & 0.046 \textcolor{ablationred}{(+2.2\%)} & 0.053 \textcolor{ablationred}{(+6.0\%)} & 0.029 (+0.0\%) \\

\bottomrule
\end{tabular}%
}
\end{table*}

\begin{figure}[h]
\centering
\includegraphics[width=0.8\linewidth]{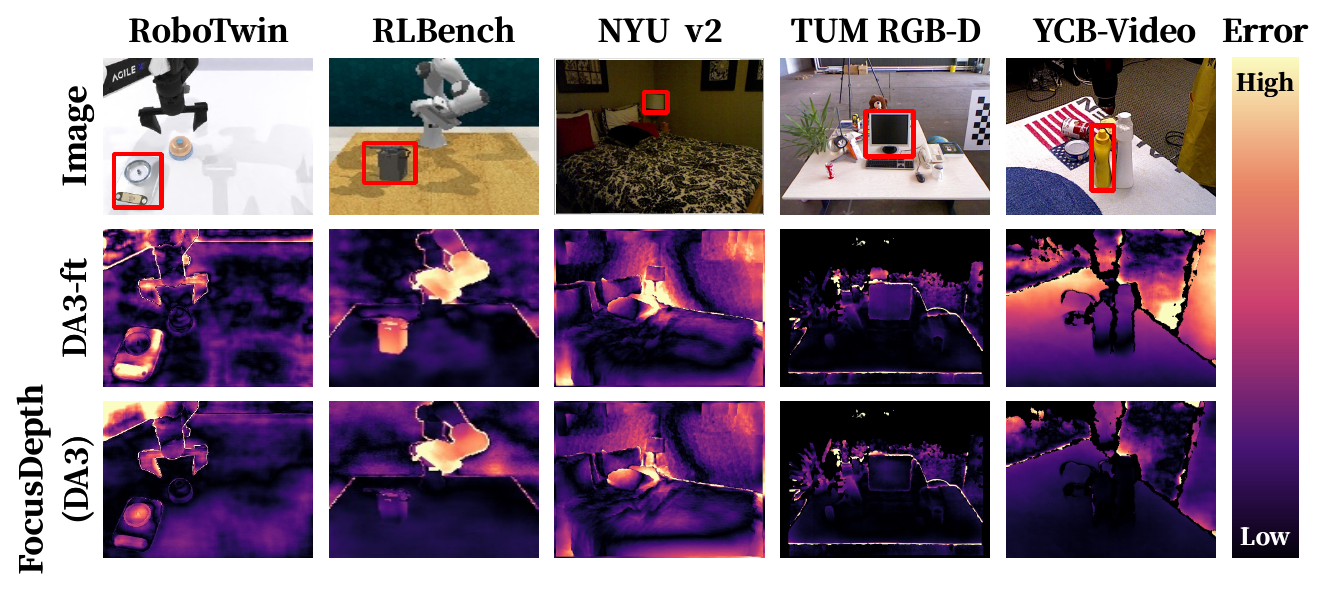}
\caption{Qualitative comparison of prompt-conditioned depth estimation. Compared with the DA3-ft, FocusDepth(DA3) produces more accurate target-region depth, cleaner boundary transitions, and more coherent local structure around the prompted object or specified target region.}
\label{fig:qualitative}
\end{figure}

\subsection{Qualitative Analysis}

Figure~\ref{fig:qualitative} shows that DA3-ft can maintain low background errors but still produces pronounced errors on complex prompted objects and their boundaries, visible as highlighted regions in the error maps. With prompt guidance, FocusDepth(DA3) substantially reduces target-region AbsRel, while largely preserving surrounding-scene depth quality. This visual evidence is consistent with the quantitative trend that FocusDepth improves local target geometry without sacrificing global scene consistency.

\section{Conclusion}

  We introduce Focusable Depth Estimation (FDE), a target-centric formulation of monocular depth estimation that prioritizes foreground accuracy and boundary
  fidelity around a specified target while preserving coherent global scene geometry. We further propose FocusDepth, a prompt-conditioned monocular relative depth estimation framework that uses
  Multi-Scale Spatial-Aligned Fusion (MSSA) to inject spatially aligned target cues into dense depth representations for conditional refinement. To make this task measurable, we establish
  FDE-Bench, which converts diverse RGB-D datasets into standardized image--target--depth triplets and evaluates depth quality over foreground, boundary, and global
  regions. Experiments on FDE-Bench show that FocusDepth improves target-region depth accuracy and boundary fidelity without compromising full-image geometric coherence,
  particularly in object-centric and manipulation-oriented scenarios. Together, these results establish FDE as a concrete task for region-prioritized depth estimation, while also indicating areas where the current study remains limited, including larger-scale joint training of FocusDepth and its application to downstream tasks, which can be explored in future work.

\clearpage
\bibliographystyle{plainnat}
\bibliography{references}

\clearpage
\appendix

\section{Technical Appendices}

\subsection{FDE-Bench}
\label{sec:app_benchmark}

\subsubsection{FDE-Bench Details}

\begin{figure}[ht]
\centering
\includegraphics[width=\linewidth]{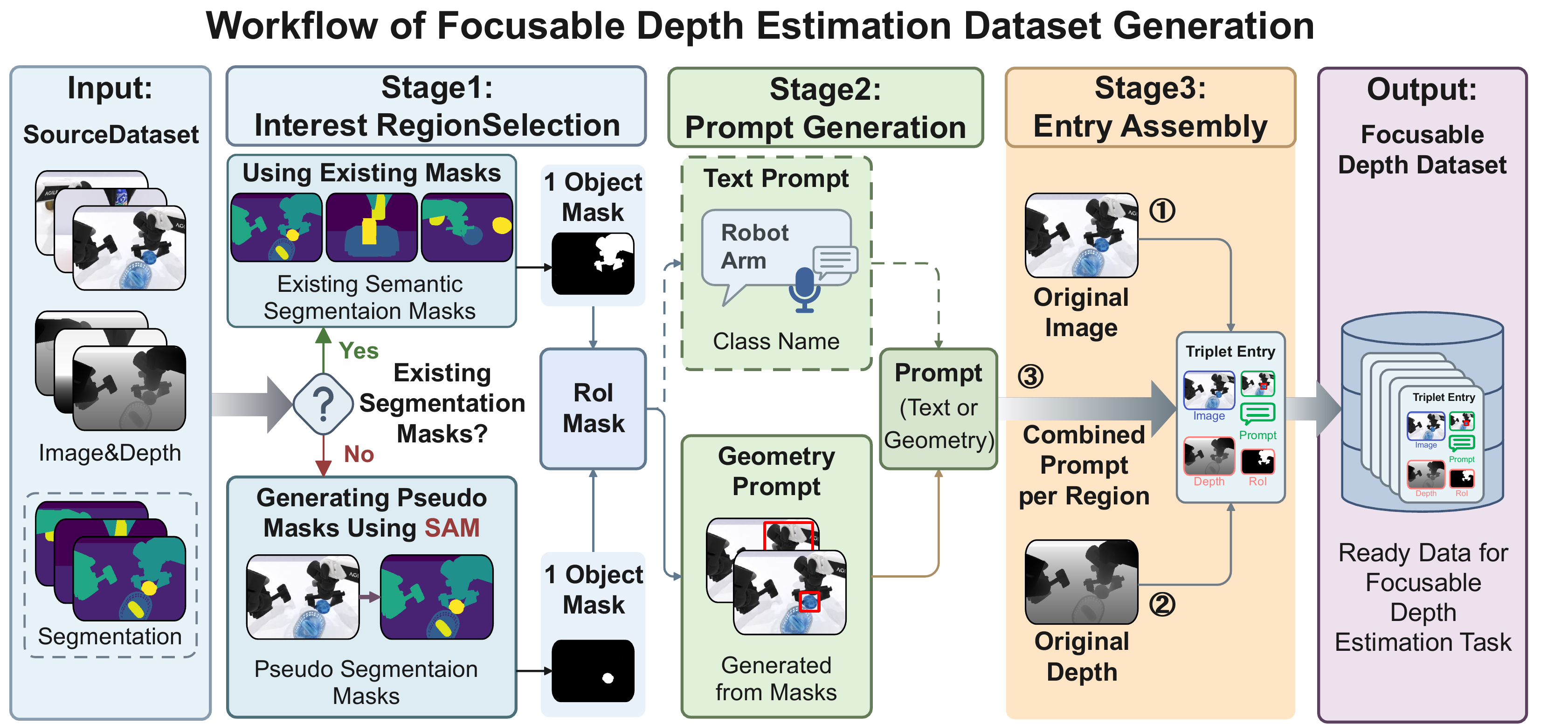}
\caption{Construction pipeline of FDE-Bench. Existing depth datasets are converted from heterogeneous annotations into standardized image--target--depth triplets for unified region-aware evaluation.}
\label{fig:benchmark-pipeline}
\end{figure}

\begin{figure}[ht]
\centering
\includegraphics[width=\linewidth]{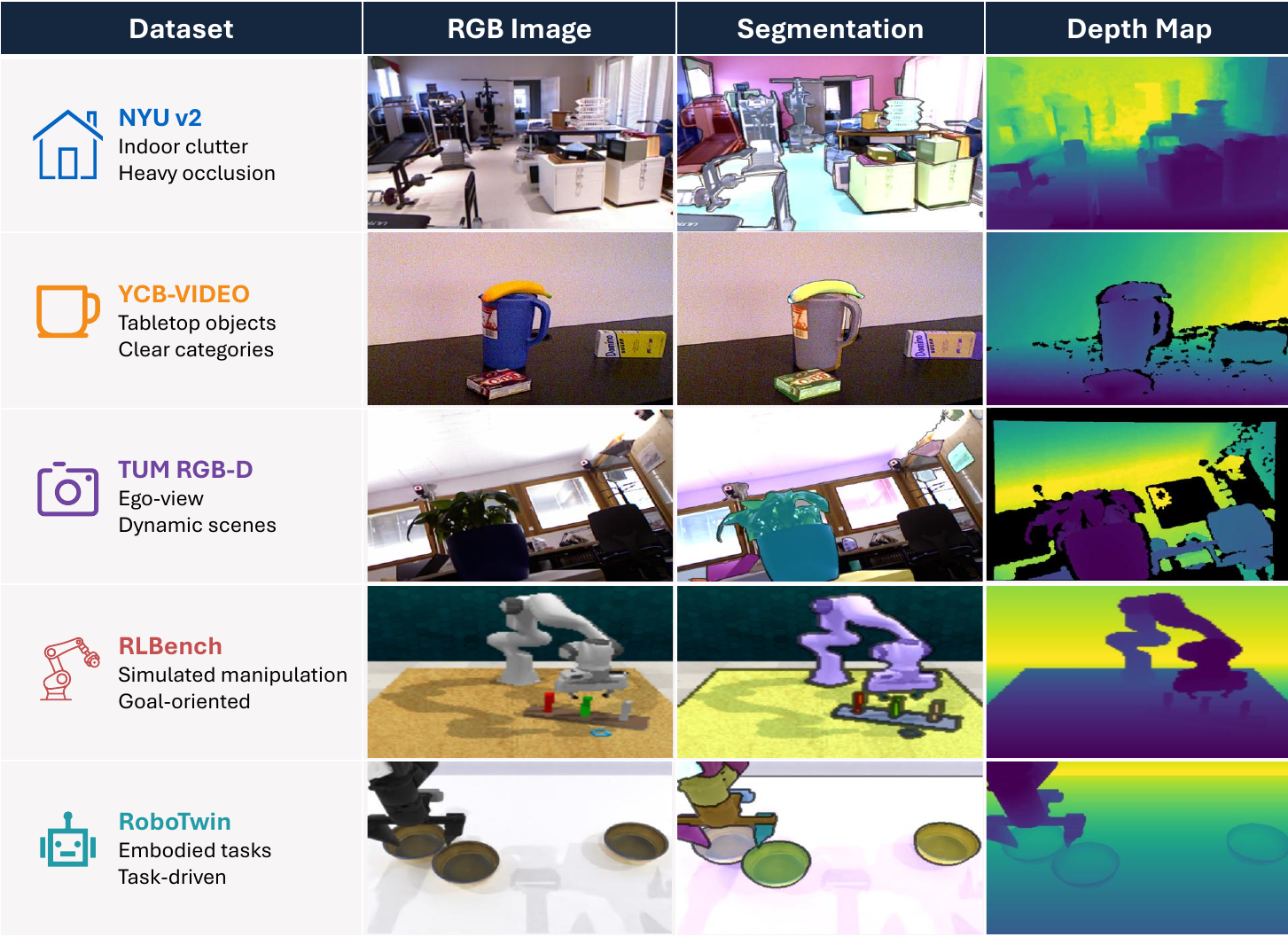}
\caption{Representative examples from the five source datasets adapted into \textbf{FDE-Bench}. For each dataset, we show an RGB image together with its segmentation and depth map, highlighting the diversity of scene structures and target-centric contexts covered by the benchmark, including indoor clutter, large-scale geometry, tabletop objects, ego-view dynamics, and embodied manipulation settings in both real and simulated environments.}
\label{appfig:dataset}
\end{figure}

\begin{figure}[ht]
\centering
\includegraphics[width=\linewidth]{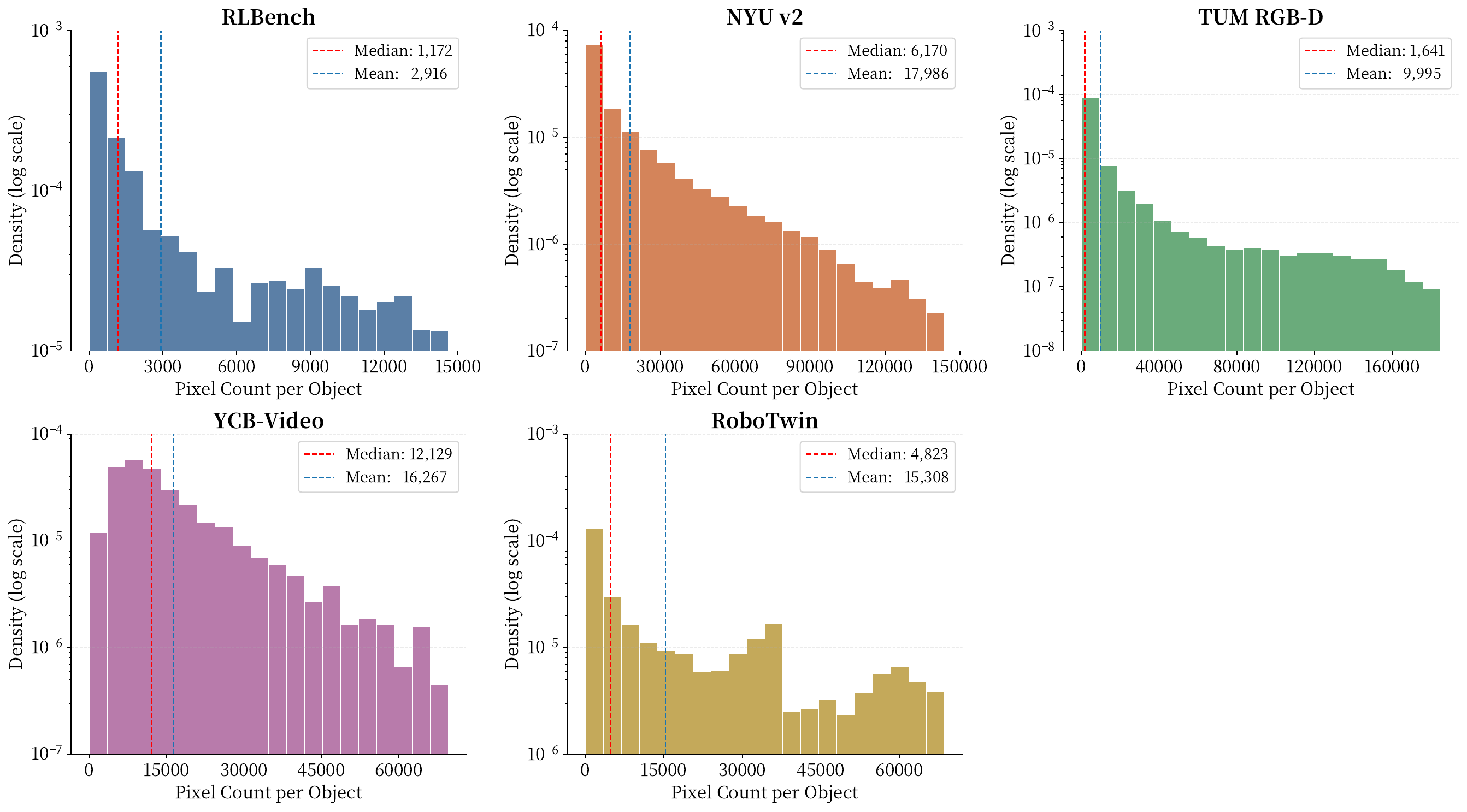}
\caption{Object size statistics across the source datasets of \textbf{FDE-Bench}, measured by pixel count per object. Each subplot shows the distribution of target-object pixel sizes for one adapted dataset, with density plotted on a logarithmic scale and the corresponding mean and median reported. The figure highlights substantial cross-dataset variation in object scale, indicating that FDE-Bench covers targets ranging from small local objects to large scene-dominant instances.}
\label{appfig:object_pixel_statistics}
\end{figure}

\begin{figure}[ht]
\centering
\includegraphics[width=\linewidth]{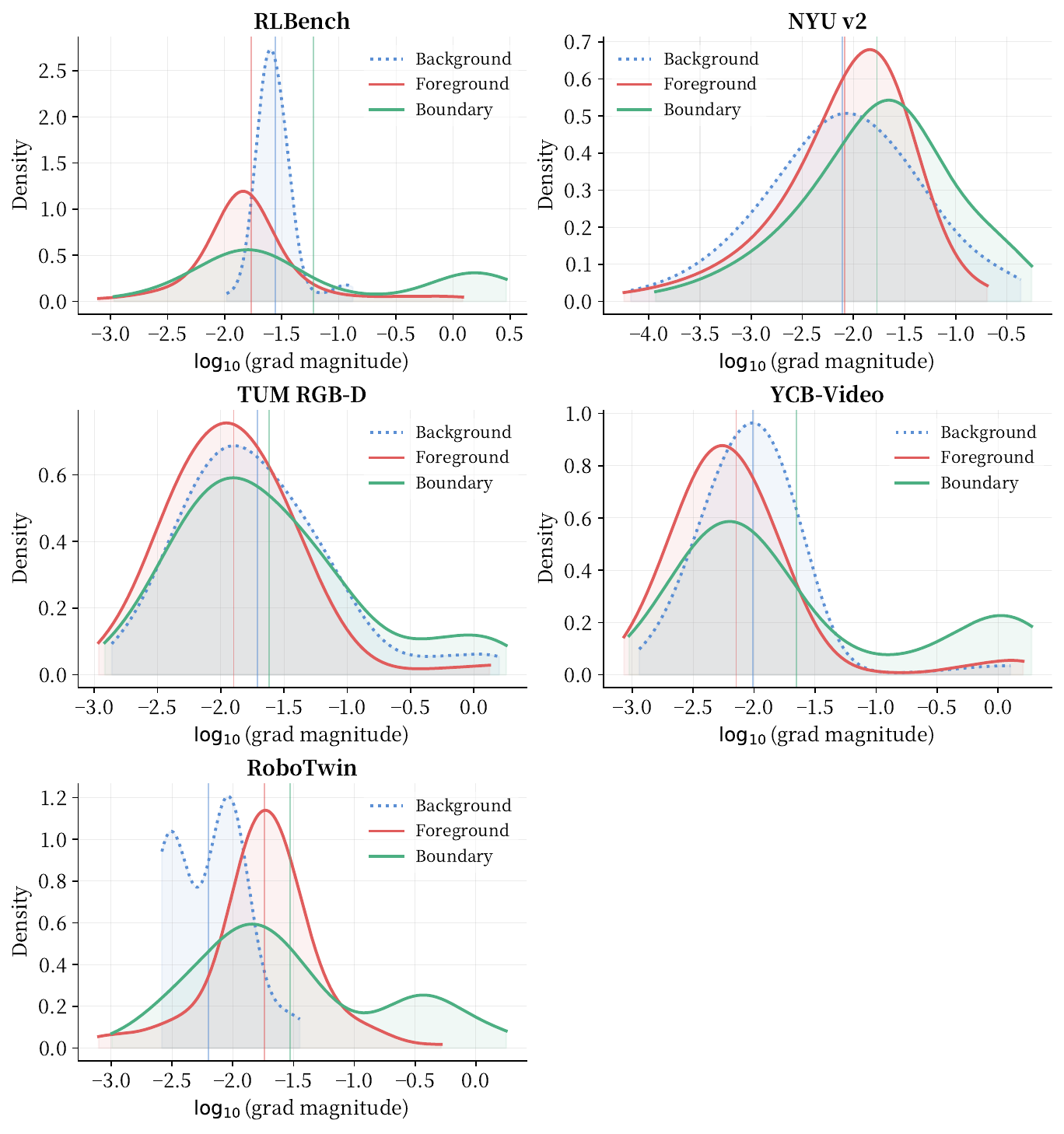}
\caption{Depth-gradient statistics across the source datasets of \textbf{FDE-Bench}. For each dataset, we plot the distribution of depth gradient magnitudes, shown as $\log_{10}(\text{grad magnitude})$, separately for background, foreground, and boundary regions. The distributions reveal systematic regional differences, with boundary areas generally exhibiting stronger depth variations, supporting the region-aware design of FDE evaluation.}
\label{appfig:grad_statistics}
\end{figure}

\begin{figure}[ht]
\centering
\includegraphics[width=\linewidth]{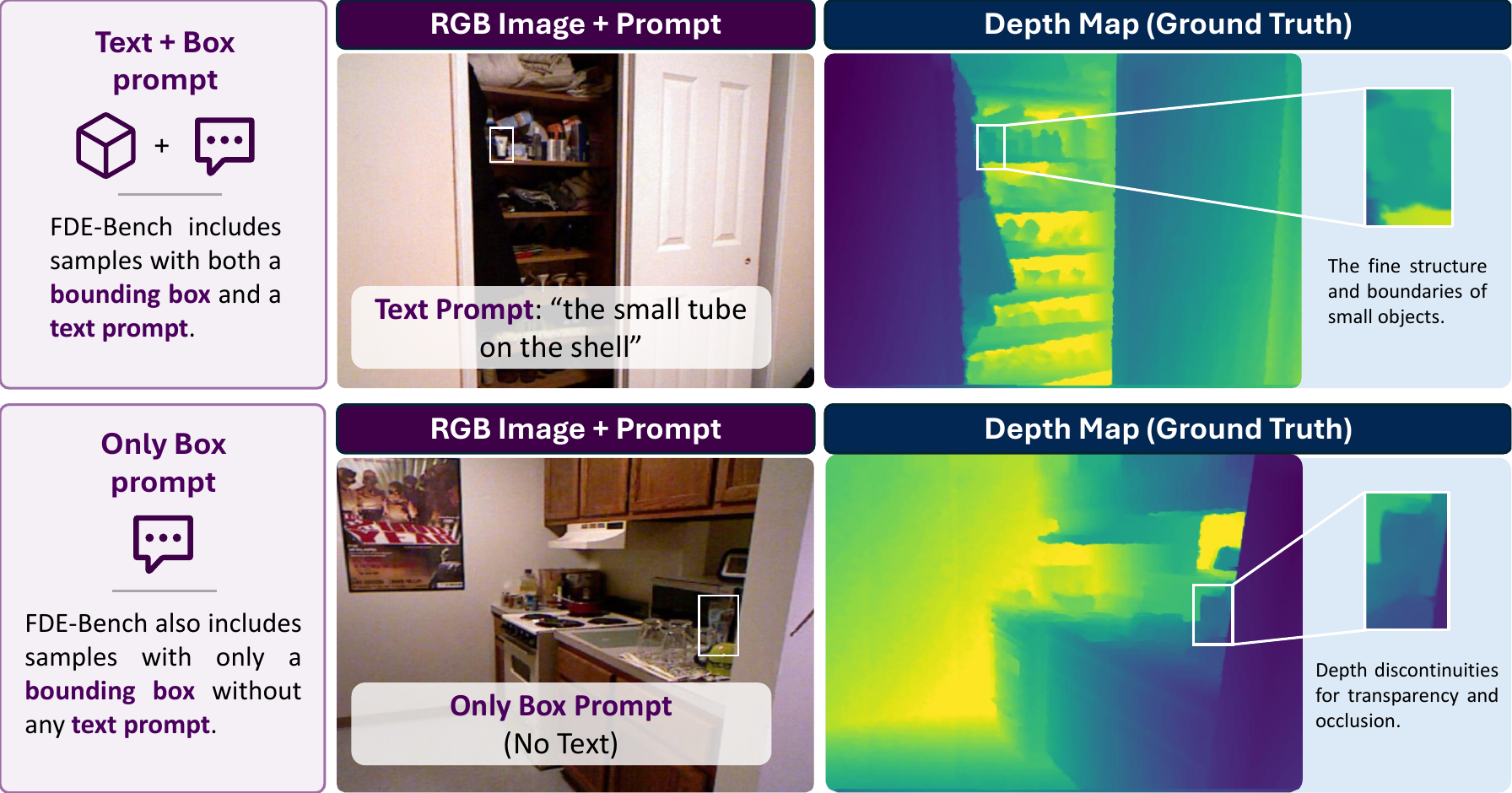}
\caption{Illustration of target specification formats in \textbf{FDE-Bench}. Depending on the adapted source dataset, a target may be specified by a bounding box together with a text prompt, or by a bounding box alone. The examples also highlight typical target-centric depth challenges addressed by FDE, including fine object structures, boundary fidelity, and depth discontinuities caused by occlusion or transparency.}

\label{appfig:prompt}
\end{figure}

This appendix complements Section~\ref{sec:benchmark} with construction details and visual statistics for \textbf{FDE-Bench}. Appendix Figure~\ref{fig:benchmark-pipeline} summarizes how heterogeneous RGB-D sources are converted into standardized image--target--depth triplets, while Appendix Figures~\ref{appfig:dataset}--\ref{appfig:prompt} document the resulting scene diversity, target-scale distribution, regional depth-gradient statistics, and supported prompt formats.

\begin{table}[h]
\centering
\scriptsize
\setlength{\tabcolsep}{3pt}
\renewcommand{\arraystretch}{0.9}
\caption{Statistics of FDE-Bench, including the images, triplets, prompt types, and object categories.}
\label{tab:benchmark-stats}
\resizebox{0.7\linewidth}{!}{%
\begin{tabular}{@{}l cc cc l c@{}}
\toprule
\multirow{2}{*}{\textbf{Dataset}} & \multicolumn{2}{c}{\textbf{Images}} & \multicolumn{2}{c}{\textbf{Triplets}} & \multirow{2}{*}{\textbf{Prompt types}} & \multirow{2}{*}{\textbf{Categories}} \\
\cmidrule(lr){2-3} \cmidrule(lr){4-5}
 & train & val & train & val & & \\
\midrule
NYU v2        & 0.8K  & 0.7K & 22.4K  & 18.4K & box / text       & 895 \\
YCB-Video     & 1.2K  & 0.1K & 7.1K   & 0.8K  & box / text       & 22  \\
TUM RGB-D     & 12.5K & 1.4K & 192.7K & 22.3K & box              & -   \\
RLBench       & 1.7K  & 1.7K & 15.1K  & 15.2K & box / text       & 55  \\
RoboTwin      & 4.3K  & 4.3K & 15.6K  & 15.8K & box              & -   \\
\midrule
\textbf{FDE-Bench} & 20.5K & 8.2K & 252.9K & 72.5K & box / text & 972 \\
\bottomrule
\end{tabular}%
}
\renewcommand{\arraystretch}{1.0}
\end{table}

The conversion pipeline retains the original RGB image and depth map from each source dataset and attaches a single target specification to each triplet. Source-provided instance or semantic masks are used when available; otherwise, SAM automatic mask generation provides pseudo masks that are filtered with the same area protocol and marked in the metadata. For each retained region, the mask is stored for evaluation, the tight bounding box is used as the box prompt, and the semantic class name is used as the text prompt when such labels are available. Consequently, one source image can produce multiple benchmark triplets that share the same RGB-D observation but differ in their target region.

\subsubsection{Dataset Scope and Statistics}

The five adapted sources listed in Table~\ref{tab:benchmark-stats} provide complementary coverage of indoor scenes, tabletop objects, ego-view motion, and manipulation environments. Appendix Figure~\ref{appfig:dataset} visualizes representative RGB, segmentation, and depth examples, and Appendix Figure~\ref{appfig:object_pixel_statistics} shows that target sizes vary substantially across sources, from small local objects to large scene-dominant regions. Prompt availability follows the annotations of each source: NYU v2, RLBench, and YCB-Video provide both box and text prompts, whereas TUM RGB-D and RoboTwin are evaluated with box prompts only, as illustrated in Appendix Figure~\ref{appfig:prompt}.

All split assignments are made at the image or sequence level to avoid leakage across target triplets. NYU v2 uses its original train/val partition, TUM RGB-D follows the benchmark-specific train/val split, and YCB-Video uses a 10\% image-level validation split. For RLBench and RoboTwin, each embodied task contributes two episodes, with one assigned to training and the other to validation.

\subsubsection{Region Definitions and Evaluation Protocol}

\label{sec:benchmark_regions}
Section~\ref{sec:benchmark} defines the foreground, boundary, and global regions used for evaluation. Here, Appendix Figure~\ref{appfig:grad_statistics} provides supporting statistics by plotting depth-gradient distributions for these regions across the source datasets. Boundary bands consistently show stronger depth variation than foreground or background areas, which motivates reporting boundary metrics separately from target-interior and whole-image metrics.

The global region is evaluated on all valid pixels rather than on a background-only complement, because scene-level geometric coherence is a whole-image property. Regional metrics are nevertheless aggregated per target: each image--target--depth triplet is scored first, and the resulting target-level values are then averaged according to the reporting protocol of each experiment. This prevents images with many retained targets from dominating the benchmark score while preserving global alignment through whole-image scale-shift normalization for relative-depth models.


\subsection{Training Hyperparameters}
\label{sec:app_training_hyperparams}

Tables~\ref{tab:focusdepth-training-hyperparams} and~\ref{tab:baseline-ft-training-hyperparams} report the training hyperparameters used for FocusDepth and the fine-tuned DA2/DA3 baselines, respectively. FocusDepth uses a two-stage schedule over the same five-epoch training budget, whereas the DA2-ft and DA3-ft baselines are trained with a single-stage fine-tuning schedule.

\begin{table}[h]
\centering
\small
\caption{Training hyperparameters for FocusDepth.}
\label{tab:focusdepth-training-hyperparams}
\begin{tabular}{lll}
\toprule
\textbf{Hyperparameter} & \textbf{Stage 1} & \textbf{Stage 2} \\
\midrule
Learning rate & $1\times10^{-4}$ & $1\times10^{-5}$ \\
Batch size per-device & 4 & 4 \\
Input resolution & $1008\times1008$ & $1008\times1008$ \\
Max steps & 1 epoch & 4 epochs \\
Warmup steps & 1/10 of Stage~1 max steps & 1/10 of Stage~2 max steps \\
LR scheduler & Cosine decay with linear warmup & Cosine decay with linear warmup \\
Weight decay & $1\times10^{-5}$ & $1\times10^{-5}$ \\
Dropout & 0.1 & 0.1 \\
Optimizer & AdamW & AdamW \\
Gradient clipping & 1.0 & 1.0 \\
\bottomrule
\end{tabular}
\end{table}

\begin{table}[h]
\centering
\small
\caption{Training hyperparameters for the DA2-ft and DA3-ft baselines.}
\label{tab:baseline-ft-training-hyperparams}
\begin{tabular}{ll}
\toprule
\textbf{Hyperparameter} & \textbf{Value} \\
\midrule
Learning rate & $1\times10^{-5}$ \\
Batch size per-device & 4 \\
Input resolution & $1008\times1008$ \\
Max steps & 5 training epochs \\
Warmup steps & 1/10 of max steps \\
LR scheduler & Cosine decay with linear warmup \\
Weight decay & $1\times10^{-5}$ \\
Dropout & 0.1 \\
Optimizer & AdamW \\
Gradient clipping & 1.0 \\
\bottomrule
\end{tabular}
\end{table}

\subsection{Additional Experimental Results}
\label{sec:app_additional_experiments}

Table~\ref{tab:main-text} reports the text-prompt comparison on NYU v2, RLBench, and YCB-Video, complementing the main box-prompt results with zero-shot baselines and fine-tuned DA2/DA3 variants.

Table~\ref{tab:finetune-box-ycb-tum-diode} reports the box-prompt comparison on TUM RGB-D and YCB-Video, where zero-shot baselines are shown together with DA3 fine-tuning variants for the datasets not included in the main box-prompt table.

\begin{table*}[t]
\centering
\small
\setlength{\tabcolsep}{3.5pt}
\caption{
Additional box-prompt results on TUM RGB-D and YCB-Video. We compare zero-shot depth foundation models, the fine-tuned DA3 baseline, and FocusDepth variants across boundary, foreground, and global regions.
}
\label{tab:finetune-box-ycb-tum-diode}
\resizebox{\textwidth}{!}{%
\begin{tabular}{llcccccc}
\toprule
\textbf{Dataset} & \textbf{Method} & \multicolumn{2}{c}{\textbf{Boundary region}} & \multicolumn{2}{c}{\textbf{Foreground region}} & \multicolumn{2}{c}{\textbf{Global region}} \\
\cmidrule(lr){3-4}\cmidrule(lr){5-6}\cmidrule(lr){7-8}
 & & $\delta_1$ $\uparrow$ & AbsRel $\downarrow$ & $\delta_1$ $\uparrow$ & AbsRel $\downarrow$ & $\delta_1$ $\uparrow$ & AbsRel $\downarrow$ \\
\midrule
\multirow{7}{*}{TUM RGB-D~\cite{sturm2012benchmark}} & UniDepth-v1~\cite{piccinelli2024unidepth} & 0.980 (0.744, 1.000) & 0.092 & \textbf{1.000} (0.867, 1.000) & 0.078 & 0.948 (0.845, 0.985) & 0.085 \\
 & UniDepth-v2~\cite{piccinelli2025unidepthv2} & 0.979 (0.805, 1.000) & \textbf{0.080} & \textbf{1.000} (\textbf{0.922}, 1.000) & \textbf{0.061} & \textbf{0.967} (\textbf{0.877}, 0.987) & \textbf{0.070} \\
 & Metric3D-v2~\cite{hu2024metric3d} & \textbf{0.984} (\textbf{0.821}, 1.000) & 0.087 & \textbf{1.000} (0.913, 1.000) & 0.076 & 0.957 (0.856, \textbf{0.989}) & 0.083 \\
 & DA2~\cite{yang2024depthv2} & 0.960 (0.709, 1.000) & 0.095 & 0.999 (0.818, 1.000) & 0.080 & 0.946 (0.849, 0.983) & 0.076 \\
\cmidrule(lr){2-8}
 & DA3~\cite{lin2025depth} & 0.964 (0.699, 1.000) & 0.099 & 0.999 (0.790, 1.000) & 0.080 & 0.938 (0.778, 0.980) & 0.093 \\
 & \cellcolor{gray!10}DA3-ft & \cellcolor{gray!10}0.999 (0.853, 1.000) & \cellcolor{gray!10}0.068 & \cellcolor{gray!10}\textbf{1.000} (0.944, 1.000) & \cellcolor{gray!10}0.057 & \cellcolor{gray!10}0.970 (0.905, 0.990) & \cellcolor{gray!10}0.060 \\
 & \cellcolor{gray!10}FocusDepth(DA3) & \cellcolor{gray!10}\textbf{1.000} (\textbf{0.882}, 1.000) & \cellcolor{gray!10}\textbf{0.061} & \cellcolor{gray!10}\textbf{1.000} (\textbf{0.966}, 1.000) & \cellcolor{gray!10}\textbf{0.051} & \cellcolor{gray!10}\textbf{0.975} (\textbf{0.912}, \textbf{0.991}) & \cellcolor{gray!10}\textbf{0.057} \\
\midrule
\multirow{7}{*}{YCB-Video~\cite{xiang2017posecnn}} & UniDepth-v1~\cite{piccinelli2024unidepth} & \textbf{1.000} (0.984, 1.000) & 0.042 & 1.000 (\textbf{0.998}, 1.000) & 0.037 & 0.993 (0.981, 0.997) & 0.045 \\
 & UniDepth-v2~\cite{piccinelli2025unidepthv2} & 0.999 (0.965, 1.000) & \textbf{0.036} & 1.000 (0.994, 1.000) & \textbf{0.026} & 0.996 (\textbf{0.991}, 0.998) & \textbf{0.031} \\
 & Metric3D-v2~\cite{hu2024metric3d} & \textbf{1.000} (\textbf{0.986}, 1.000) & 0.040 & 1.000 (\textbf{0.998}, 1.000) & 0.031 & \textbf{0.997} (0.988, \textbf{0.999}) & 0.042 \\
 & DA2~\cite{yang2024depthv2} & 0.999 (0.963, 1.000) & 0.045 & 1.000 (0.993, 1.000) & 0.038 & 0.994 (0.972, 0.998) & 0.040 \\
\cmidrule(lr){2-8}
 & DA3~\cite{lin2025depth} & 0.994 (0.938, 1.000) & 0.054 & 0.999 (0.986, 1.000) & 0.045 & 0.992 (0.961, 0.997) & 0.047 \\
 & \cellcolor{gray!10}DA3-ft & \cellcolor{gray!10}\textbf{1.000} (0.988, 1.000) & \cellcolor{gray!10}0.042 & \cellcolor{gray!10}\textbf{1.000} (\textbf{0.999}, 1.000) & \cellcolor{gray!10}0.036 & \cellcolor{gray!10}0.997 (0.984, \textbf{0.999}) & \cellcolor{gray!10}0.044 \\
 & \cellcolor{gray!10}FocusDepth(DA3) & \cellcolor{gray!10}\textbf{1.000} (\textbf{0.990}, 1.000) & \cellcolor{gray!10}\textbf{0.037} & \cellcolor{gray!10}\textbf{1.000} (\textbf{0.999}, 1.000) & \cellcolor{gray!10}\textbf{0.026} & \cellcolor{gray!10}\textbf{0.998} (\textbf{0.992}, \textbf{0.999}) & \cellcolor{gray!10}\textbf{0.032} \\
\bottomrule
\end{tabular}%
}
\end{table*}

\subsection{RLBench Prompt-Correctness Mean--Median Comparison}
\label{sec:app_prompt_correctness}

The prompt-correctness study in Section~\ref{sec:experiment_prompt_correctness} reports mean values, unlike the main comparative results that report medians. We use means in this diagnostic experiment because prompt perturbations can create bad cases whose aggregate effect is important for measuring robustness. Table~\ref{tab:app-prompt-correctness-mean-median} reports both means and medians for the same RLBench class-specific text-prompt setting, showing how the diagnostic mean values relate to the median-based protocol used in the main comparison.

\begin{table*}[t]
\centering
\scriptsize
\setlength{\tabcolsep}{2.5pt}
\caption{Mean--median comparison for the RLBench prompt-correctness study under class-specific text prompts. The main paper reports mean values for this diagnostic study to better reflect bad cases under prompt perturbations; medians are included here for comparison with the main evaluation metrics.}
\label{tab:app-prompt-correctness-mean-median}
\resizebox{\textwidth}{!}{%
\begin{tabular}{lcccccccccccc}
\toprule
Method & \multicolumn{4}{c}{Boundary region} & \multicolumn{4}{c}{Foreground region} & \multicolumn{4}{c}{Global region} \\
\cmidrule(lr){2-5}\cmidrule(lr){6-9}\cmidrule(lr){10-13}
 & $\delta_1$ mean $\uparrow$ & $\delta_1$ med. $\uparrow$ & AbsRel mean $\downarrow$ & AbsRel med. $\downarrow$
 & $\delta_1$ mean $\uparrow$ & $\delta_1$ med. $\uparrow$ & AbsRel mean $\downarrow$ & AbsRel med. $\downarrow$
 & $\delta_1$ mean $\uparrow$ & $\delta_1$ med. $\uparrow$ & AbsRel mean $\downarrow$ & AbsRel med. $\downarrow$ \\
\midrule
Correct prompt & \textbf{0.919} & \textbf{0.995} & \textbf{0.108} & \textbf{0.045} & \textbf{0.901} & \textbf{1.000} & \textbf{0.137} & \textbf{0.050} & \textbf{0.943} & 0.996 & 0.075 & \textbf{0.029} \\
Wrong prompt (other visible object) & 0.914 & 0.995 & 0.111 & 0.045 & 0.884 & \textbf{1.000} & 0.148 & 0.052 & \textbf{0.943} & \textbf{0.996} & \textbf{0.074} & \textbf{0.029} \\
Wrong prompt (absent object) & 0.911 & 0.994 & 0.117 & 0.047 & 0.886 & \textbf{1.000} & 0.152 & 0.053 & 0.939 & 0.995 & 0.078 & 0.030 \\
Empty prompt & 0.910 & 0.993 & 0.118 & 0.045 & 0.886 & \textbf{1.000} & 0.154 & 0.051 & 0.939 & 0.995 & 0.078 & \textbf{0.029} \\
DA3-ft no-prompt baseline & 0.860 & 0.973 & 0.162 & 0.068 & 0.804 & 0.998 & 0.226 & 0.088 & 0.915 & 0.984 & 0.105 & 0.041 \\
\bottomrule
\end{tabular}%
}
\end{table*}

\begin{figure}[h]
\centering
\includegraphics[width=\linewidth]{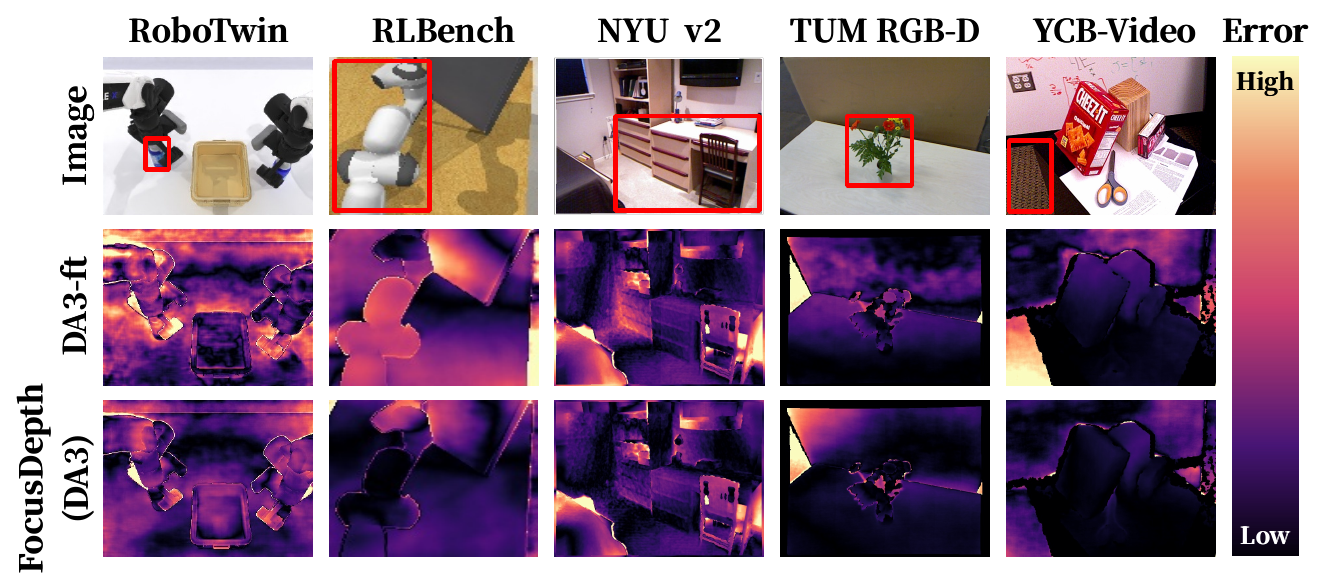}
\caption{Additional qualitative comparison of prompt-conditioned depth estimation. Compared with the DA3-ft, FocusDepth(DA3) produces more accurate target-region depth, cleaner boundary transitions, and more coherent local structure around the prompted object or specified target region.}
\label{fig:app_qualitative}
\end{figure}

\begin{figure}[h]
\centering
\includegraphics[width=\linewidth]{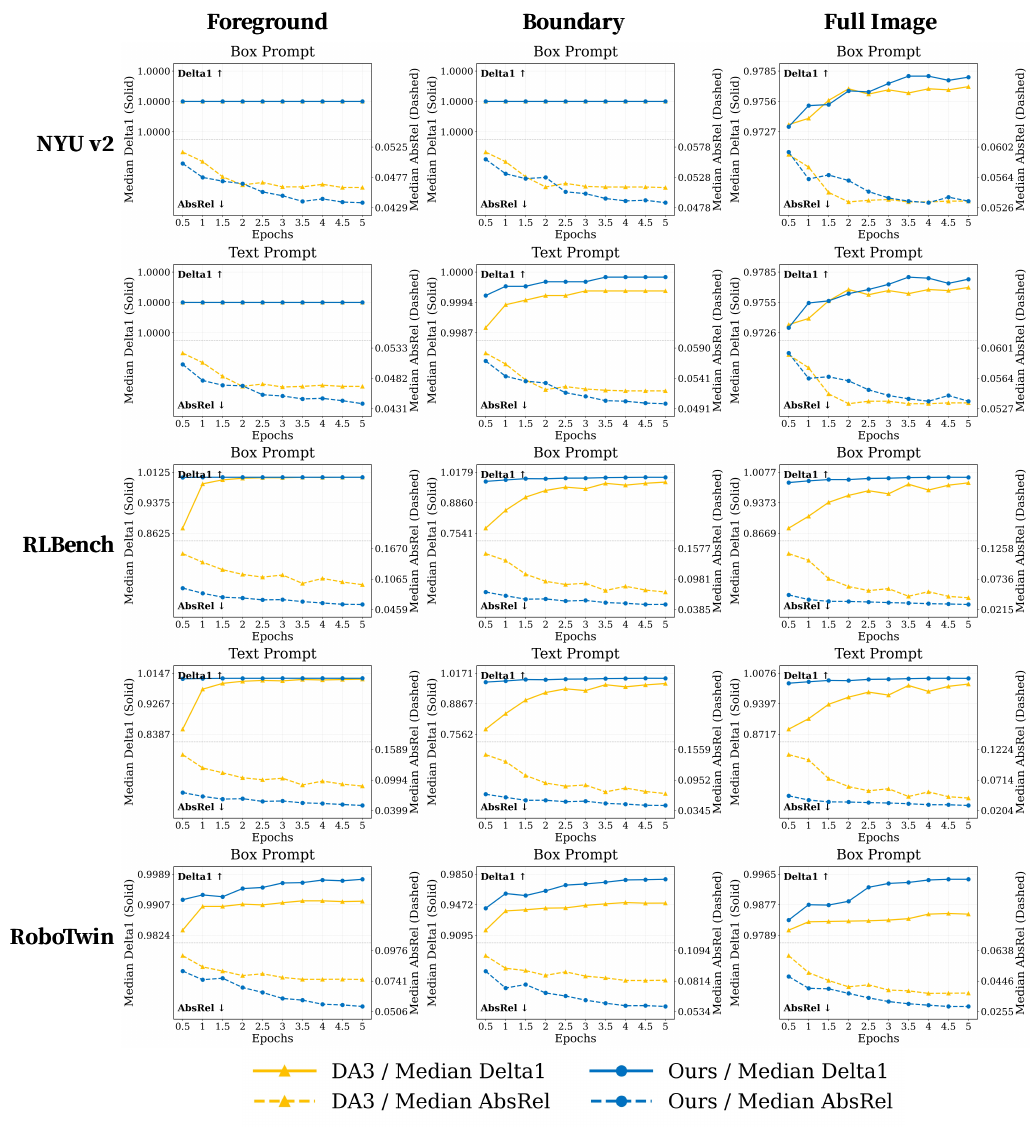}
\caption{
Validation curves of DA3-ft and FocusDepth(DA3) over training epochs. 
We compare $\delta_1$ and AbsRel on NYU v2, RLBench, and RoboTwin across foreground, boundary, and global regions. 
FocusDepth(DA3) consistently improves target-relevant regions, especially foreground and boundary areas, while preserving global depth quality.
}
\label{fig:app_training_curve}
\end{figure}



\end{document}